\documentclass[journal]{IEEEtran}
\usepackage{amsmath,amsfonts}
\usepackage{array}
\usepackage[caption=false,font=normalsize,labelfont=sf,textfont=sf]{subfig}
\usepackage{textcomp}
\usepackage{stfloats}
\usepackage{url}
\usepackage{verbatim}
\usepackage{algpseudocode}
\usepackage{amsmath}
\usepackage[nocompress]{cite}
\usepackage{booktabs}
\usepackage{multirow}
\usepackage{siunitx}
\usepackage{graphicx}
\usepackage{makecell}
\usepackage[colorlinks=true,
            linkcolor=blue,
            citecolor=blue,
            urlcolor=blue]{hyperref}
\usepackage{lipsum}
\usepackage{setspace}
\usepackage[ruled,vlined,linesnumbered]{algorithm2e}

\makeatletter

\newcommand{\Rmnum}[1]{\expandafter\@slowromancap\romannumeral #1@}
\makeatother
\begin{document}

\title{\fontsize{22pt}{27pt}\selectfont
Advancing Visual Reliability: Color-Accurate Underwater Image Enhancement for Real-Time Underwater Missions\vspace{0.2em}}

\author{Yiqiang Zhou$^{\dagger}$, Yifan Chen$^{\dagger}$, Zhe Sun$^{\dagger, \ast}$, Jijun Lu, Ye Zheng,  and Xuelong Li$^{\ast}$,~\IEEEmembership{Fellow,~IEEE}
\vspace{-0.5cm}
\thanks{
This work has been submitted to the IEEE for possible publication. Copyright may be transferred without notice, after which this version may no longer be accessible.

Yiqiang Zhou is with the Institute of Artificial Intelligence (TeleAI), China Telecom, P. R. China, and also with the College of Computer and Control Engineering, Northeast Forestry University, Harbin, 150040, China. This work was done during his research internship at TeleAI (e-mail: cloudyu1215@outlook.com).

Yifan Chen is with the Institute of Artificial Intelligence (TeleAI), China Telecom, P. R. China, and also with the College of Future Information Technology, Fudan University, Shanghai 200433, China (e-mail: chenyifan@fudan.edu.cn).

Zhe Sun is with the Institute of Artificial Intelligence (TeleAI), China Telecom, P. R. China, also with the School of Artificial Intelligence, Optics and Electronics (iOPEN), Northwestern Polytechnical University, Xi'an 710072, China, and also with the Fujian Ocean Innovation Center, Xiamen 361102, China (e-mail: sunzhe@nwpu.edu.cn).

Jijun Lu, Ye Zheng, and Xuelong Li are with the Institute of Artificial Intelligence (TeleAI), China Telecom, P. R. China (e-mail: jijun\_lu@163.com; zhengye@westlake.edu.cn; xuelong\_li@ieee.org).

$^{\dagger}$ These authors contributed equally.

$^{\ast}$ Corresponding authors: Zhe Sun, and Xuelong Li.

$^{\scalebox{0.6}{$\spadesuit$}}$ Code is available at \textbf{\url{https://github.com/Cloudyu1215/UIE}}.
}

}

\markboth{ }%
{Shell \MakeLowercase{\textit{et al.}}: A Sample Article Using IEEEtran.cls for IEEE Journals}


\maketitle

\begin{abstract}

Underwater image enhancement plays a crucial role in providing reliable visual information for underwater platforms, since strong absorption and scattering in water-related environments generally lead to image quality degradation. Existing high-performance methods often rely on complex architectures, which hinder deployment on underwater devices. Lightweight methods often sacrifice quality for speed and struggle to handle severely degraded underwater images. To address this limitation, we present a real-time underwater image enhancement framework with accurate color restoration. First, an Adaptive Weighted Channel Compensation module is introduced to achieve dynamic color recovery of the red and blue channels using the green channel as a reference anchor. Second, we design a Multi-branch Re-parameterized Dilated Convolution that employs multi-branch fusion during training and structural re-parameterization during inference, enabling large receptive field representation with low computational overhead. Finally, a Statistical Global Color Adjustment module is employed to optimize overall color performance based on statistical priors. Extensive experiments on eight datasets demonstrate that the proposed method achieves state-of-the-art performance across seven evaluation metrics. The model contains only 3,880 inference parameters and achieves an inference speed of 409 FPS. Our method improves the UCIQE score by 29.7\% under diverse environmental conditions, and the deployment on ROV platforms and performance gains in downstream tasks further validate its superiority for real-time underwater missions. 

\end{abstract}


\vspace{-1em}
\section{Introduction}
\IEEEPARstart{U}{nderwater} image enhancement (UIE) is widely applied in water-related engineering, providing essential visual support for diverse underwater missions, including search-and-rescue, archaeological exploration, and infrastructure inspection\cite{peng2017underwater,PhotoniX,zhang2022underwater}.
However, constrained by the distinctive light absorption and scattering properties of water, captured images usually suffer from severe color distortion and quality degradation\cite{zongshu01,single}. Current methods often encounter a trade-off between enhancement quality and computational performance during edge deployment: high-performance models struggle to meet the real-time requirements of embedded systems due to high computational costs, while lightweight solutions are compelled to sacrifice color quality for speed. Consequently, achieving both low resource consumption and high-quality color restoration for underwater terminals and downstream visual tasks has become a critical challenge.

Considerable efforts are devoted to color restoration and lightweight architectural design\cite{non_learn}. The underwater light propagation mechanism differs from that in atmospheric environments, exhibiting severe absorption and random scattering. On one hand, long-wavelength red light attenuates rapidly as depth increases, causing captured images to typically appear blue-green and distorting the true colors of objects. On the other hand, light scattering induced by suspended particles causes contrast degradation and detail blurring. Early studies primarily rely on physical models or priors to model degradation\cite{darkchannelpiror,transmission,carlevaris,redchannel,support1}, or leverage information from the less attenuated green channel to compensate for the heavily degraded red channel\cite{redbuchang}. Although these traditional methods mitigate color shift, their compensation weights are often  set to fixed values manually, failing to cover dynamic underwater scenes with varying turbidity and depths, leading to over or under-compensation in complex lighting environments.

Recently, learning-based methods make significant progress in UIE tasks. Researchers have utilized architectures such as Convolutional Neural Networks (CNNs), 
Generative Adversarial Networks (GANs), Transformers, and Mamba\cite{an2024uwmamba,watergan,tianyufeng} to achieve better visual quality for degraded images through end-to-end training. However, mainstream models often adopt complex architectures like attention mechanisms or Diffusion models\cite{UIEdm} for peak performance. But their massive computational costs result in high latency on underwater platforms, hindering time-sensitive tasks. While unified lightweight restoration frameworks are developed for various degradation scenarios, their efficacy in underwater environments remains limited. Existing models fail to address the coupled underwater optical degradation effects, often leading to uneven color and missing texture details. Thus, it remains difficult for lightweight models to maintain real-time performance without compromising the visual fidelity required for underwater missions.

\begin{figure}[t]
\centering
\includegraphics[width=\columnwidth]{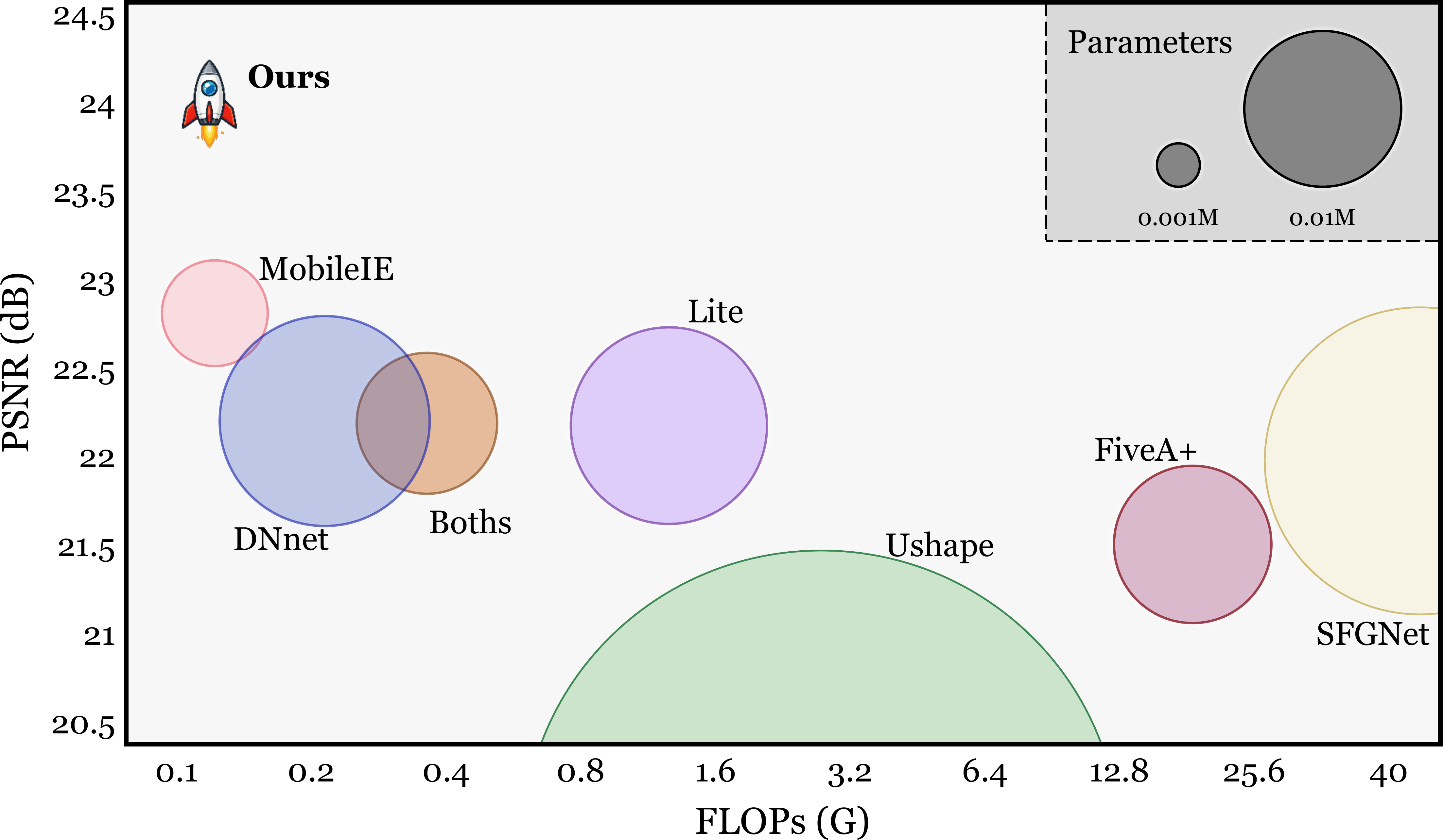}
\caption{Comparison of different methods in the FLOPs–PSNR space on UIEB dataset. The x-axis denotes FLOPs (G, 256 × 256 inputs), y-axis indicates PSNR (dB), and circle size is used to indicate parameter count.}
\label{14}
\end{figure}

To overcome these challenges, we propose a lightweight UIE framework that balances efficient inference with color restoration. Rather than relying solely on increased network capacity, we construct an enhancement pipeline consisting of three collaborative modules: channel compensation, detail enhancement, and global correction. First, we design the Adaptive Weighted Channel Compensation (AWCC) module, which uses the relatively stable green channel as a reference anchor and employs adaptive weights to dynamically compensate for the damaged red and blue channels. Second, to restore structures and textures without increasing computational cost, we propose Multi-branch Re-parameterized Dilated Convolution (MRDConv), which captures diverse receptive field information via parallel branches during training and collapses into a single convolution for inference. Finally, we further introduce a Statistical Global Color Adjustment (SGCA) module that learns and predicts color temperature, tint, and saturation adjustment parameters based on statistical priors. As shown in Fig.~\ref{14}, these designs allow the model to achieve a balance between authentic color restoration and real-time inference with extremely low parameters and complexity. 

Our main contributions are as follows:

\begin{itemize}
    \item We present a lightweight UIE framework centered on high-fidelity color restoration, achieving high-quality visual enhancement in resource-constrained scenarios to provide stable inputs for underwater terminals.
    \item We introduce an AWCC strategy for dynamic color restoration, and further develop the MRDConv and SGCA modules for texture enhancement and global color adjustment, respectively, enabling high-fidelity underwater image enhancement.
    \item Extensive evaluations on eight datasets show that our method reaches state-of-the-art performance across seven metrics, while requiring only 3,880 inference parameters and running at 409 FPS on GPU. Compared with current advanced methods, it uses 4.13\% fewer parameters and improves PSNR by 1.2 dB.
    \item In underwater camera applications, our method achieves a 29.7\% improvement in the UCIQE metric. Furthermore, deployment on ROV platforms, along with performance gains in downstream instance segmentation, further validate its superiority for real-time underwater missions.
\end{itemize}

\vspace{2em}

\section{Related Work}
Existing UIE methods have evolved from traditional approaches to deep learning-based frameworks, and more recently to lightweight architectures for efficient deployment. The following subsections review these representative paradigms and their respective advantages and limitations.
\subsection{Traditional Methods}
Traditional UIE methods primarily rely on physical imaging models. One category is based on underwater optical propagation mechanisms, restoring scene radiance by modeling wavelength-dependent attenuation and particle scattering\cite{tcsvt4}. For instance, methods based on the Dark Channel Prior (DCP) and its extensions\cite{darkchannelpiror,prior2,prior3} introduce color correction while compensating for haze-like effects. Drews Jr et al.\cite{transmission} propose the Underwater Dark Channel Prior (UDCP) assuming significant red channel attenuation, while Carlevaris-Bianco et al.\cite{carlevaris} utilize the attenuation difference between RGB channels to predict transmission. While these methods improve naturalness, they often depend on accurate depth, water type, or lighting parameters, limiting their generalization in complex environments.

Another category employs enhancement and fusion strategies without explicit physical assumptions\cite{lichongyi}. Retinex-based methods enhance local contrast by separating illumination and reflectance; UMSHE\cite{UMSHE} alleviates global brightness shifts via block-based histogram equalization; Ancuti et al.\cite{fusion} fuses white-balanced and original images with weight maps to enhance edges and color contrast; and WFAC\cite{WFAC} utilizes waveform decomposition fusion to improve quality. Despite their low computational overhead, these methods are limited to over-saturation, color distortion, or artifacts under complex degradation, highlighting the limitations of handcrafted priors.

\subsection{Deep Learning-based Methods}
With the rapid advancement of deep learning, UIE research has shifted toward data-driven end-to-end frameworks\cite{tcsvt,lixuelong03}. Many works utilize CNNs to build models\cite{cnn02,smdr-si}. For example, UWCNN\cite{UWCNN} explores underwater scene priors with a lightweight structure; Water-Net\cite{waternet} implements adaptive fusion of multiple inputs via learned confidence maps; and WWE-UIE\cite{wweuie} introduces an adaptive white balance module within a lightweight CNN to improve efficiency.

Transformers have been widely applied to underwater image enhancement for modeling long-range dependencies. UWAGA\cite{UWAGA} introduces a grouped attention mechanism built upon Swin Transformer. U-shape Transformer\cite{U} enhances representation ability through multi-scale feature fusion and global context modeling. Spectroformer\cite{Spectroformer} performs joint transmission–illumination modeling using a multi-domain query cascade. Nevertheless, these approaches still show limited robustness under severely degraded conditions. To further address this issue, methods such as PhaseFormer\cite{Phaseformer} incorporate Fourier Transform to explicitly handle low-frequency color distortion and high-frequency detail degradation.

Diffusion models further framed UIE as a conditional denoising process\cite{Lixuelong02}. UIE-DM\cite{UIEdm} first applies conditional diffusion to UIE to balance performance and efficiency; WF-Diff\cite{wfdiff} integrates Fourier priors into diffusion. However, the above high-complexity models are unsuitable for real-time deployment on edge devices during underwater missions.

\begin{figure*}
\centering
\includegraphics[width=\textwidth]{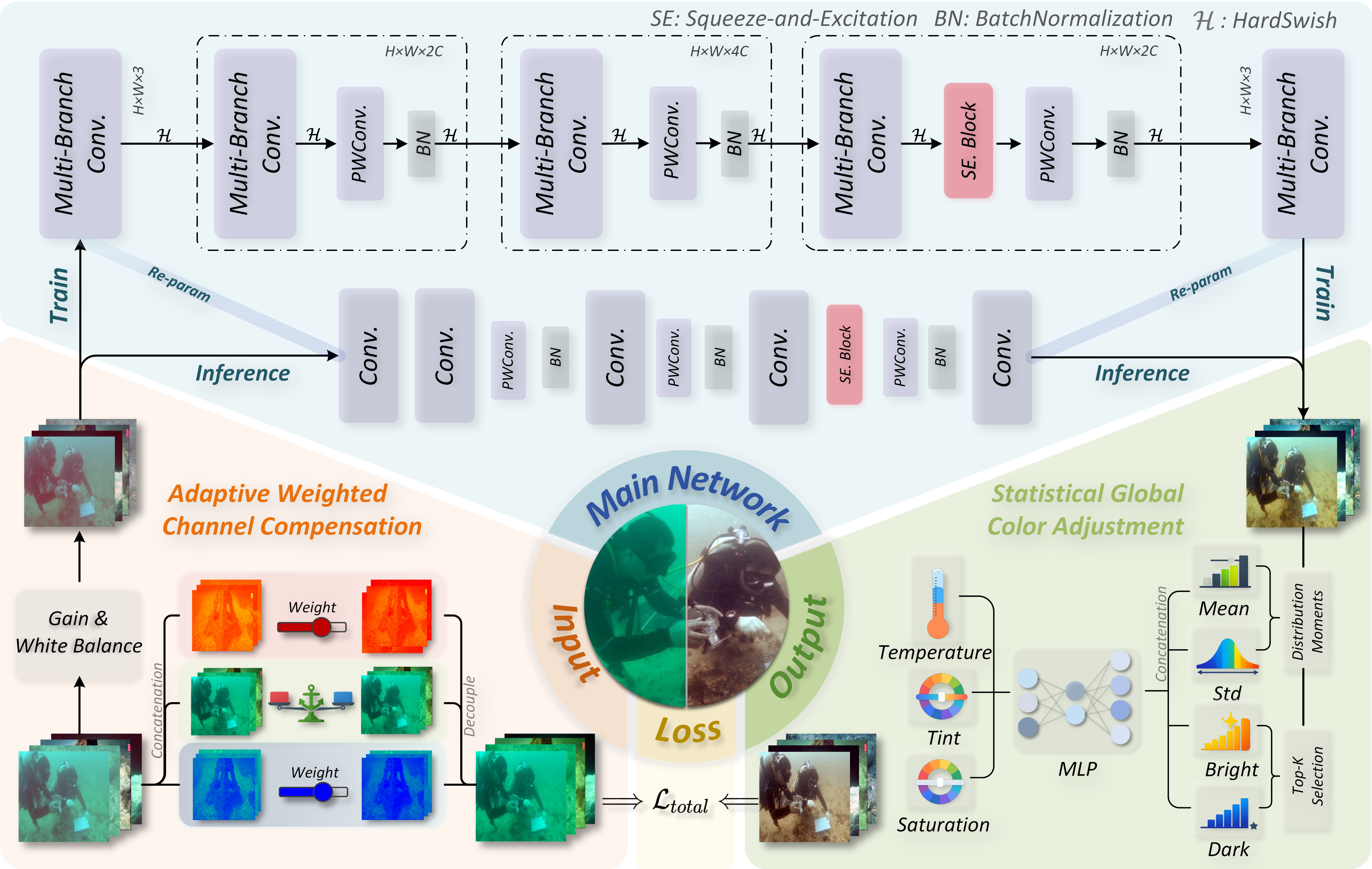}
\caption{Overview of our proposed UIE framework. The pipeline begins with the AWCC module, which adaptively compensates degraded color channels and stabilizes luminance through secondary correction. The enhanced image is then processed by a CNN backbone incorporating MRDConv to capture  multi-directional features with efficient inference. Finally, the SGCA module leverages global statistical priors to refine color temperature, tint, and saturation, producing high-fidelity enhancement with low computational overhead.}
\label{pipeline}
\end{figure*}

\subsection{Lightweight Methods}
To meet real-time deployment requirements, lightweight designs have become key research areas. Foundational work focused on efficient operators, such as MobileNet\cite{Mobilenetv3} with depthwise separable convolutions, ShuffleNet\cite{shufflenet} with channel shuffling, and GhostNet\cite{Ghost} or FasterNet\cite{chen2023run} which enhance feature interaction efficiency through cost-effective operations. Recently, research has pivoted toward structural re-parameterization and hybrid architectures. Models like FastViT\cite{fastvit} and RepViT\cite{repvit} decouple training complexity from inference efficiency. MobileIE\cite{MobileIE} specifically adopts mobile-friendly operators for the restoration task.

Several models are proposed to handle the constraints of limited power and complex optical degradation\cite{cnn01,puie}. DNnet\cite{dnnet} balances restoration quality and parameter count via compact connections. FiveA+\cite{5a} optimizes the balance between network depth and functional modules to maintain high throughput. SFGNet\cite{sfg} and LiteEnhanceNet\cite{lite} utilize efficient feature fusion and depthwise separable convolutions to mitigate color shifts with minimal overhead. Boths\cite{boths} explores structural designs like SD-D and SD-I3 to improve robustness across water types.

Despite these advancements, a persistent contradiction remains between extreme efficiency and high color fidelity. Many lightweight UIE models achieve high frame rates through aggressive pruning, which often leads to non-uniform color or detail loss in complex scenes. Therefore, developing a framework capable of high-quality color restoration and structural preservation under a strict computational budget remains a critical challenge.

\section{Method}
This section presents the proposed UIE framework, which is designed to jointly enhance color fidelity and deployment efficiency through a unified pipeline composed of three collaborative modules. The overall design is described in the following subsections.
\subsection{Overall Pipeline}

The overall architecture of the UIE framework is illustrated in Fig.~\ref{pipeline}. In the first stage, the Adaptive Weighted Channel Compensation (AWCC) module serves as a pre-processing step. It uses the relatively well-preserved green channel as an anchor to dynamically compensate for the severely attenuated red and blue channels, followed by a secondary correction by adopting the Gray World Assumption to stabilize luminance. Next, the main network adopts a CNN architecture, within which Multi-branch Re-parameterized Dilated Convolution (MRDConv) is embedded as a core component. By using structural re-parameterization, it utilizes multi-scale receptive fields to capture multi-directional information during training, while ensuring efficient inference through a simplified single-path structure. Finally, the Statistical Global Color Adjustment (SGCA) module extracts four global statistical priors to refine color temperature, tint, and saturation via a lightweight Multi-Layer Perceptron (MLP). Through the synergistic design, this pipeline achieves high-fidelity restoration with minimal computational overhead.

\subsection{Adaptive Weighted Channel Compensation}

Due to the wavelength-dependent attenuation characteristics of underwater optical propagation, underwater images commonly suffer from severe color distortion and low contrast. Among visible light, red light has the longest wavelength and attenuates fastest in water. As a result, underwater scenes often exhibit a blue-green color shift. In contrast, the green channel generally suffers less attenuation in most water types\cite{redbuchang,support1} and thus retains more color information.
\begin{figure}
\centering
\includegraphics[width=\columnwidth]{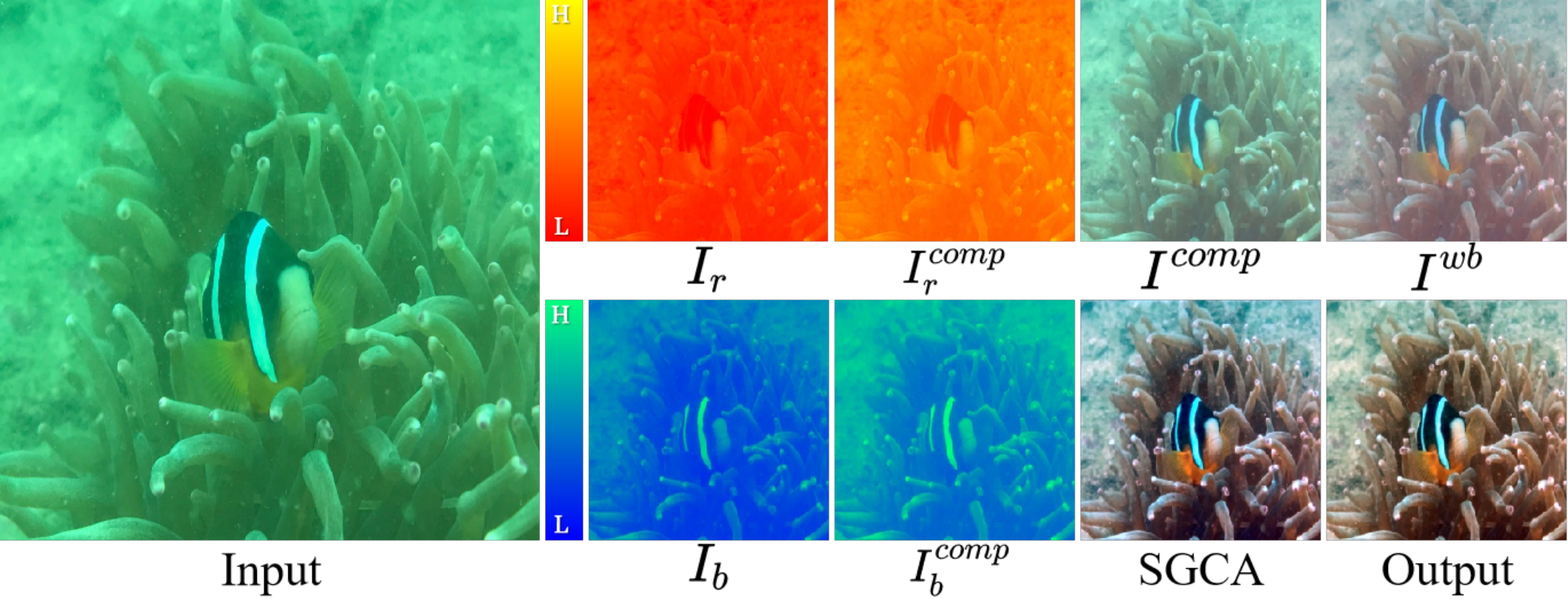}
\caption{Visualization of intermediate values for AWCC and the SGCA modules: The red and blue channels are visualized using the AUTUMN and WINTER colormaps, respectively.}
\label{02_vis}
\end{figure}

Motivated by these observations, we propose an Adaptive Weighted Channel Compensation module. The main idea is to use the relatively stable green channel as a guiding anchor for compensating the heavily degraded red and blue channels through two learnable weights, enabling scene-adaptive color restoration. Given a degraded input image $I\in \mathbb{R} ^{H\times W\times 3}$ , the mean value of each channel is first computed:
\begin{equation}
I_k^{mean} = \frac{1}{HW} \sum_{i=1}^H \sum_{j=1}^W I_k(i,j), \quad k \in \{r, g, b\}
\end{equation}

The green channel is then used to adaptively refine the red and blue channels. The compensated red and blue outputs, denoted by $I_r^{comp}$ and $I_b^{comp}$, are formulated as:
\begin{equation}
I_r^{comp} = I_r + \alpha_r \cdot (I_g^{mean} - I_r^{mean})
\end{equation}
\begin{equation}
I_b^{comp} = I_b + \alpha_b \cdot (I_g^{mean} - I_b^{mean})
\end{equation}
where $\alpha_r$ and $\alpha_b$ are learnable adjustment factors that are dynamically optimized through back propagation. This allows the model to adapt to diverse color degradation patterns across varying underwater environments and imaging conditions. Fig.~\ref{02_vis} illustrates the visualization of intermediate feature maps in the channel compensation process, along with the outputs of the SGCA module described in the following section.

However, while channel compensation effectively mitigates color shift, it often introduces over-exposure. To overcome this limitation, we further perform a secondary correction mechanism based on the Gray World Assumption. This refinement step operates on the premise that, under ideal white balance conditions, the spatial means of the compensated image $I^{comp}$ channels should converge toward a common, uniform gray value $\mu_{gray}$:
\begin{equation}
I^{comp}=Concat[I_r^{comp},I_g,I_b^{comp}]
\end{equation}
\begin{equation}
\mu_{gray} = \frac{1}{3} \sum_{c \in \{r,g,b\}} M_c
\end{equation}
where $Concat(\cdot)$ denotes channel-wise concatenation, $M_c$ denotes the average irradiance estimate for each channel of $I^{comp}$. We then estimate per-channel gains $G_c$ and apply linear scaling to achieve the final white-balanced output $I^{wb}$:
\begin{equation}
G_c = \frac{\mu_{gray}}{M_c + \epsilon}
\end{equation}
\begin{equation}
I^{wb} = I^{comp} \odot G_c
\end{equation}
where $\epsilon = 1 \times 10^{-6}$ is added as a small constant to prevent zero-division issues. This strategy achieves robust color reconstruction even in extreme attenuation environments.

\begin{figure}[!t]
\centering
\includegraphics[width=0.85\columnwidth]{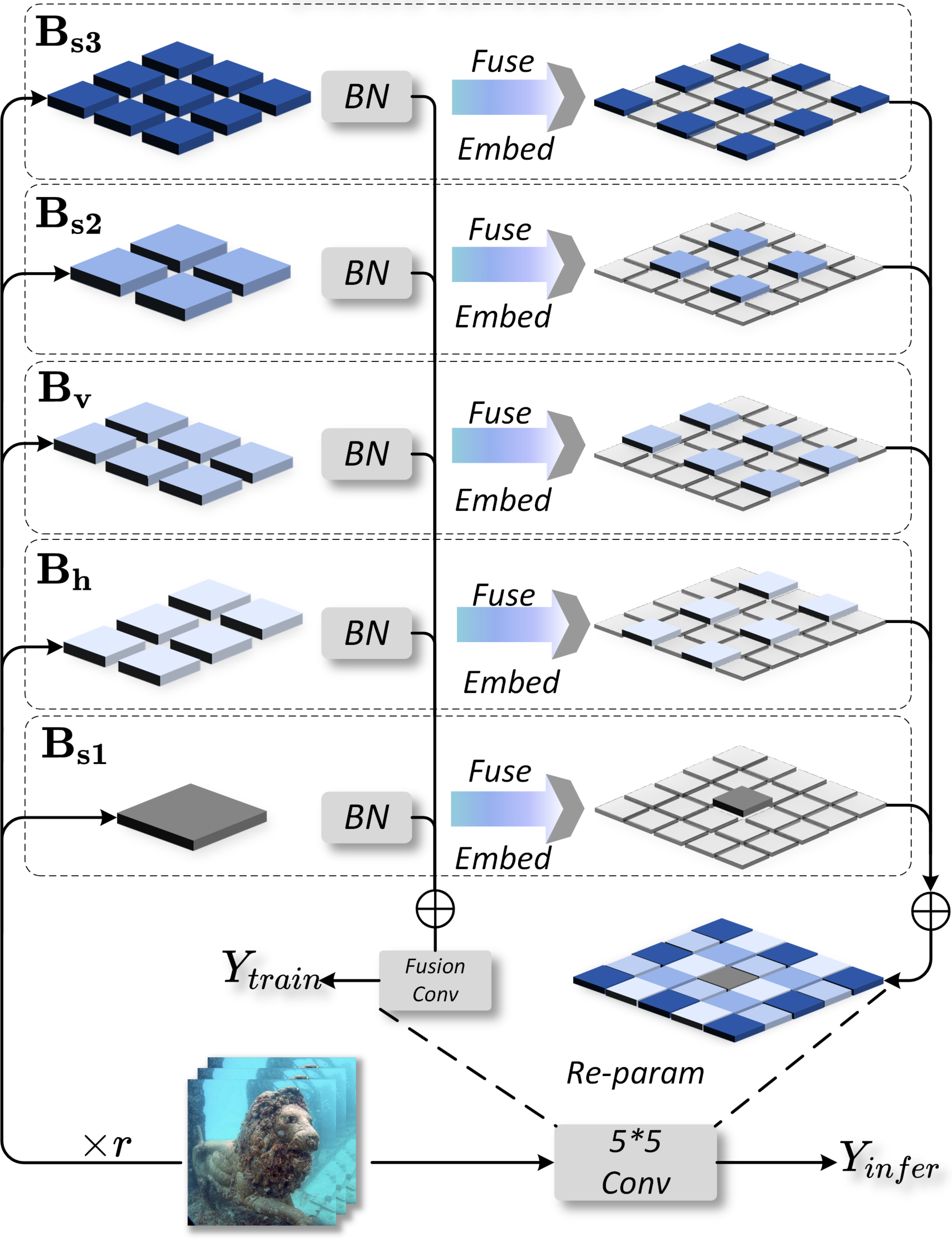}
\caption{Architecture of the MRDConv and the Re-parameterization process.}
\label{MRDConv}
\end{figure}

\subsection{Multi-branch Re-parameterized Dilated Convolution}
In CNNs, expanding the receptive field is vital for capturing global contextual information. However, simply applying large, dense kernels (e.g., $7 \times 7$ or $11 \times 11$) comes with significant computational burden and a large parameter count. To enlarge the receptive field and enhance feature extraction capabilities without significantly increasing inference costs, we propose a dilated convolution module based on structural re-parameterization, denoted as Multi-branch Re-parameterized Dilated Convolution (MRDConv).

The module receives an input tensor $X\in \mathbb{R}^{C_{in}\times H\times W } $ and processes it through five parallel dilated convolutional branches as illustrated in Fig.~\ref{MRDConv}. These branches are designed to simulate a large dense receptive field while maintaining a low parameter count. Specifically, the five branches are defined as follows:

$\mathbf{B_{s3}} $: $3 \times 3$ kernel with $dilation=2, padding=2$, resulting in an effective receptive field of $5 \times 5$.

$\mathbf{B_{s2}} $: $2 \times 2$ kernel with $dilation=2$, capturing asymmetric local features.

$\mathbf{B_{s1}} $: $1 \times 1$ kernel designed to preserve and project identity-like features across the expanded channel dimension.

$\mathbf{B_{v}} $: $3 \times 2$ kernel with a $dilation=(2, 2),padding=(2, 1)$, for capturing vertical information.

$\mathbf{B_{h}} $: $2 \times 3$ kernel with a $dilation=(2, 2),padding=(1, 2)$, for capturing horizontal information.

To improve representational capacity, the middle channels $C_{mid}$ for all branches are expanded by a factor of Rep-scale, denoted as $r$ relative to the output channels $C_{out}$ ($r=4$ in this paper). Each branch consists of a convolution followed by Batch Normalization to stabilize optimization and improve conditioning. The outputs of the five branches are fused via element-wise addition and then projected back to the target output dimension $C_{out}$ through a $1 \times 1$ fusion layer. The forward pass during training is summarized in Algorithm~\ref{algo1}:

\begin{algorithm}[ht]
\caption{Phase Training}
\label{algo1}
\SetAlgoLined
\setstretch{1.2}
\KwIn{$\mathbf{X} \in \mathbb{R}^{C_{in}\times H\times W}$}

$C_{mid}\leftarrow C_{out}\times r$\;
\ForEach{$i\in\{s3,s2,s1,v,h\}$}{
    $\mathbf{X}_i \leftarrow \mathrm{BN}_i(\mathrm{Conv}_i(\mathbf{X}))$\;
}
$\mathbf{X}_{sum}\leftarrow \sum_i \mathbf{X}_i$\;
$\mathbf{Y}\leftarrow \mathrm{FusionConv}_{1\times1}(\mathbf{X}_{sum})$\;
\KwOut{$\mathbf{Y} \in \mathbb{R}^{C_{out}\times H\times W}$}
\end{algorithm}

Although the multi-branch architecture benefits training, it increases inference latency. To address this, MRDConv adopts structural re-parameterization to collapse all branches into a single $5 \times 5$ convolution during inference. As shown in Fig.~\ref{MRDConv} and Algorithm~\ref{algo2}, the overall process can be divided into three steps: fusing the convolution and BN parameters of each branch into equivalent weights and biases, embedding the resulting sparse kernels into a zero-initialized $5 \times 5$ grid and summing them into a dense kernel, and finally absorbing the terminal $1 \times 1$ fusion layer by channel-wise matrix multiplication to obtain $W_{final}$ and $b_{final}$. In this way, MRDConv preserves the advantages of multi-branch training and large receptive fields while maintaining efficient single-path inference.

\begin{algorithm}[t]
\caption{Phase Re-parameterization}
\label{algo2}
\setstretch{1.2}
\SetAlgoLined
\KwIn{Trained branch parameters $\{(\mathrm{Conv}_i,\mathrm{BN}_i)\}_{i\in\{s3,s2,s1,v,h\}}$ and fusion parameters $(\mathbf{W}_f,\mathbf{b}_f)$}

\ForEach{$i\in\{s3,s2,s1,v,h\}$}{
    $(\mathbf{W}'_i,\mathbf{b}'_i)\leftarrow \mathrm{Fuse}(\mathrm{Conv}_i,\mathrm{BN}_i)$\;
    $\mathbf{K}_i \leftarrow \mathrm{Embed}_{5\times5}(\mathbf{W}'_i;\, i)$ 
}
$\mathbf{K}_{sum}\leftarrow \sum_i \mathbf{K}_i$\;
$\mathbf{b}_{sum}\leftarrow \sum_i \mathbf{b}'_i$\;

$\mathbf{W}_{final}\leftarrow \mathrm{Reshape}\!\left(\mathbf{W}_f^{flat}\cdot \mathbf{K}_{sum}^{flat}\right)$\;
$\mathbf{b}_{final}\leftarrow \mathbf{W}_f^{flat}\mathbf{b}_{sum}+\mathbf{b}_f$\;
\KwOut{Equivalent $5*5 \ Conv$ parameters $(\mathbf{W}_{final},\mathbf{b}_{final})$}
\end{algorithm}

\subsection{Statistical Global Color Adjustment}

The color shift and contrast degradation commonly observed in underwater images are primarily governed by the overall water environment and illumination conditions, rather than local structural variations. Based on this observation, we avoid computationally expensive per-pixel mappings and instead describe the underwater scene using four complementary types of global statistical features. A set of parameters is then predicted to perform global color adjustment.

$\mu_c  \And  \sigma_c$: We calculate two distribution moments, the channel-wise mean and standard deviation to reflect the overall brightness variation and contrast range.

$v_{bright}$: Bright-region statistics used to estimate the dominant illumination color of the scene. For each channel, the mean intensity of the top 5\% brightest pixels is computed.

$v_{dark}$: Dark-region statistic used to model the haze component in underwater imaging caused by back scattering, which is typically concentrated in the darkest regions. For each channel, the mean intensity of the bottom 5\% pixels is computed.

The final statistical prior is defined as:
\begin{equation}
V_{stat} = Concat[\mu_c, \sigma_c, v_{bright}, v_{dark}]
\end{equation}

Then, $V_{stat}$ is fed into a lightweight MLP to predict three global scalars: temperature shift $\Delta T$, tint shift $\Delta \tau$, and saturation gain $S_{gain}$:

\begin{equation}
\Delta T, \Delta \tau = \lambda_t \cdot \tanh(MLP(V_{stat}))
\end{equation}
\begin{equation}
S_{gain} = 1 + \lambda_s \cdot \tanh(MLP(V_{stat}))
\end{equation}
where $\lambda_t = 0.15$ and $\lambda_s = 0.5$ are hyperparameters used to constrain excessive shifts. Then, the color adjustment is applied in RGB space:

\begin{equation}
\begin{aligned}
X'_R &= X_R + \Delta T, \\
X'_B &= X_B - \Delta T, \\
X'_G &= X_G - \Delta \tau.
\end{aligned}
\end{equation}

Saturation is subsequently adjusted using the Rec.709 standard luminance map ($Y$) to preserve structural integrity.
\begin{equation}
X_{out}=Y+S_{gain}\cdot (X'-Y)
\end{equation}

By integrating the statistical priors, the SGCA module effectively refines the final output. This lightweight adjustment allows the model to handle diverse underwater environments without the heavy computational cost of pixel-wise global modeling.

\begin{table*}[t]
\centering
\caption{Quantitative comparison of performance and computational efficiency on UIEB, LSUI, EUVP-S, EUVP-D and EUVP-I datasets. Best results are in \textbf{bold}.}
\label{tab:main_results}
\resizebox{\textwidth}{!}{
\begin{tabular}{l l cccc cccc cccc}
\toprule
\multirow{2}{*}{Method} & \multirow{2}{*}{Venue} &
\multicolumn{4}{c}{UIEB} &
\multicolumn{4}{c}{LSUI} &
\multicolumn{4}{c}{EUVP-S} \\
\cmidrule(lr){3-6}\cmidrule(lr){7-10}\cmidrule(lr){11-14}
& & PSNR$\uparrow$ & SSIM$\uparrow$ & LPIPS$\downarrow$ & UIQM$\uparrow$ &
PSNR$\uparrow$ & SSIM$\uparrow$ & LPIPS$\downarrow$ & UIQM$\uparrow$ &
PSNR$\uparrow$ & SSIM$\uparrow$ & LPIPS$\downarrow$ & UIQM$\uparrow$ \\
\midrule
U-Shape\cite{U}  & IEEETIP'23   & 20.518 & 0.782 & 0.215 & \textbf{3.280} & 24.492 & 0.861 & 0.148 & 3.119 & 23.503 & 0.827 & 0.187 & 3.130 \\
FiveA+\cite{5a}  & BMVC'23      & 21.566 & 0.892 & 0.134 & 3.092 & 23.883 & 0.901 & 0.164 & 2.993 & 25.968 & \textbf{0.887} & 0.159 & 2.932 \\
Boths\cite{boths}& IEEEGRSL'23  & 22.230 & 0.901 & 0.156 & 3.273 & 21.221 & 0.851 & 0.222 & 3.083 & 23.487 & 0.838 & 0.268 & 3.038 \\
SFGNet\cite{sfg} & ICASSP'24    & 22.100 & 0.882 & 0.140 & 3.010 & 24.641 & 0.899 & 0.162 & 2.967 & 25.815 & 0.866 & 0.178 & 3.037 \\
Lite\cite{lite}  & ESWA'24      & 22.227 & 0.894 & 0.130 & 3.129 & 20.596 & 0.825 & 0.237 & 2.937 & 25.139 & 0.842 & 0.190 & 2.946 \\
DNnet\cite{dnnet}& ESWA'25      & 22.247 & 0.884 & 0.126 & 2.870 & 22.774 & 0.897 & 0.201 & 2.842 & 21.873 & 0.859 & 0.245 & 2.830 \\
Phaseformer\cite{Phaseformer}& WACV'25      & 23.469 & 0.842 & \textbf{0.106} & 3.279 & 25.431 & 0.831 & 0.150 & \textbf{3.146} & 24.132 & 0.746 & 0.200 & \textbf{3.138} \\
MobileIE\cite{MobileIE}      & ICCV'25      & 22.840 & 0.894 & 0.123 & 3.019 & 25.294 & 0.898 & 0.142 & 2.989 & 26.890 & 0.882 & 0.158 & 2.946 \\
\midrule
Ours           & -& \textbf{24.055} & \textbf{0.902} & 0.121 & 3.045 & \textbf{26.296} & \textbf{0.907} & \textbf{0.133} & 3.029 & \textbf{27.435} & 0.878 & \textbf{0.126} & 3.053 \\
\bottomrule
\end{tabular}
}

\vspace{4pt}

\resizebox{\textwidth}{!}{
\begin{tabular}{l l cccc cccc cccc}
\toprule
\multirow{2}{*}{Method} & \multirow{2}{*}{Venue} &
\multicolumn{4}{c}{EUVP-D} &
\multicolumn{4}{c}{EUVP-I} & FPS & Params & Size & FLOPs
\\
\cmidrule(lr){3-6}\cmidrule(lr){7-10}\cmidrule(lr){11-11}\cmidrule(lr){12-12}\cmidrule(lr){13-13}\cmidrule(lr){14-14}
& & PSNR$\uparrow$ & SSIM$\uparrow$ & LPIPS$\downarrow$ & UIQM$\uparrow$ &
PSNR$\uparrow$ & SSIM$\uparrow$ & LPIPS$\downarrow$ & UIQM$\uparrow$ & $640\times480$ & (K) & (MB) & (G)
\\
\midrule

U-Shape\cite{U}        & IEEETIP'23  & 21.199 & 0.868 & 0.258 & 3.210 & 23.902 & 0.844 & 0.198 & 3.097 & 25.984   & 22817  & 87.04  & 2.983 \\
FiveA+ \cite{5a}        & BMVC'23     & 21.571 & 0.894 & 0.270 & 3.082 & 24.137 & 0.857 & 0.195 & 3.109 & 90.224   & 8.974  & 0.03   & 18.74 \\
Boths\cite{boths}          & IEEEGRSL'23 & 21.025 & 0.879 & 0.250 & 3.108 & 23.008 & 0.843 & 0.206 & 3.156 & 389.431  & 6.447  & 0.02   & 0.4256 \\
SFGNet\cite{sfg}         & ICASSP'24   & 22.235 & 0.899 & 0.241 & 3.127 & 24.627 & 0.870 & 0.198 & 3.140 & 23.82    & 1298   & 4.95   & 40.787  \\
Lite\cite{lite} & ESWA'24     & 21.763 & \textbf{0.903} & 0.252 & 3.064 & 23.898 & 0.830 & 0.202 & 3.087 & 150.289  & 13.688 & 0.05   & 1.38 \\
DNnet\cite{dnnet}          & ESWA'25     & 19.877 & 0.874 & 0.265 & 3.088 & 22.635 & 0.856 & 0.216 & 3.089 & 1636.25  & 15.51  & 0.24   & 0.251 \\
Phaseformer\cite{Phaseformer}    & WACV'25     & 21.424 & 0.877 & 0.258 & 3.177 & 23.525 & 0.801 & 0.195 & \textbf{3.232} & 45.89    & 1,779  & 6.8    & 13.041 \\
MobileIE\cite{MobileIE}       & ICCV'25     & 22.304 & 0.892 & \textbf{0.238} & 3.138 & 24.835 & 0.860 & 0.207 & 3.109 & 678.5    & 4.047  & 0.02   & 0.146 \\

\midrule
Ours           & -          & \textbf{22.336} & 0.899 & 0.243 & \textbf{3.233} & \textbf{24.853} & \textbf{0.875} & \textbf{0.194} & 3.205 & 409.09 & \textbf{3.88} & \textbf{0.01} & \textbf{0.145} \\
\bottomrule
\end{tabular}
}
\end{table*}

\subsection{Loss Function}

To jointly optimize structural fidelity, perceptual quality, and color accuracy, we adopt a composite loss consisting of pixel-level, perceptual, and color-alignment constraints. Denote the enhanced image and the ground-truth by $I_{out}$ and $I_{gt}$, respectively. The total loss is defined as:

\textbf{Charbonnier Loss.}
A robust differentiable variant of $L_1$ loss is used to preserve structural details while reducing sensitivity to outliers:
\begin{equation}
\mathcal{L}_{char} = \frac{1}{N} \sum_{i=1}^{N} \sqrt{(I_{out, i} - I_{gt, i})^2 + \epsilon^2}
\end{equation}
where $N$ is the number of pixels and $\epsilon = 1 \times 10^{-8}$.

\textbf{PSNR Loss.}
To directly optimize reconstruction quality, a normalized PSNR-based loss is employed:
\begin{equation}
\mathrm{RMSE} = \sqrt{\frac{1}{CHW} \sum_{c,h,w} (I_{out} - I_{gt})^2 + \epsilon}
\end{equation}

\begin{equation}
\mathcal{L}_{psnr} = \frac{50 - 20 \cdot \log_{10} \left( \frac{1}{\mathrm{RMSE}} \right)}{100}
\end{equation}

\textbf{Perceptual Loss.}
To enhance perceptual quality, we adopt a VGG-19 based perceptual loss that measures feature-level similarity:
\begin{equation}
\mathcal{L}_{vgg} = \| \phi(I_{out}) - \phi(I_{gt}) \|_2^2
\end{equation}
where $\phi(\cdot)$ denotes features extracted from the VGG-19 network.

\textbf{Color Consistency Loss.}
To improve color fidelity, we constrain the angular distance between RGB vectors of the enhanced image and the reference:
\begin{equation}
\mathcal{L}_{color} = \frac{1}{HW} \sum_{p} \arccos \left( \frac{I_{out, p} \cdot I_{gt, p}}{\|I_{out, p}\| \|I_{gt, p}\| + \epsilon} \right)
\end{equation}

The final loss is a weighted combination of all terms:
\begin{equation}
\mathcal{L}_{total} = \mathcal{L}_{char}+2 \cdot \mathcal{L}_{psnr} + 0.01 \cdot \mathcal{L}_{vgg} + \mathcal{L}_{color}
\end{equation}

\section{Experiment}
In this section, the proposed method is evaluated from three perspectives: overall performance, ablation analysis, and downstream underwater task performance. Quantitative and qualitative experiments are first conducted, followed by ablation studies and robustness validation on downstream tasks.
\subsection{Underwater Image Enhancement}
\subsubsection{Settings}
The proposed method is compared with eight representative approaches. For quantitative evaluation, both training and testing are performed on the UIEB\cite{waternet}, LSUI\cite{U}, and EUVP\cite{cnn01} (contains three subsets: Dark, ImageNet, and Scenes) datasets. These datasets contain 890(UIEB), 4,279(LSUI), 2,185(EUVP-D), 5,500(EUVP-I), and 3,700(EUVP-S) paired images, respectively. An 8:2 ratio is used to divide all datasets into training and testing sets, with all images resized to $256 \times 256$. For qualitative evaluation, the model pre-trained on UIEB and LSUI is used for inference on U45\cite{u45}, RUIE\cite{ruie}, and ColorCheck7\cite{redbuchang} to assess its generalization ability.

To evaluate restoration quality, we utilize Peak Signal-to-Noise Ratio(PSNR), Structural Similarity Index(SSIM)\cite{quality}, and Learned Perceptual Image Patch Similarity(LPIPS) as metrics for full-reference datasets. For non-reference datasets, Underwater Image Quality Measure(UIQM)\cite{Uiqm} and Underwater Color Image Quality Evaluation(UCIQE)\cite{Uciqe} are used to comprehensively assess color richness, contrast, and overall perceptual quality. Additionally, we adopt the CIEDE2000\cite{ciede2000} metric on the ColorCheck7 dataset to quantify color restoration accuracy via perceptual color difference.

All experiments are conducted using the PyTorch framework on a single NVIDIA A100 GPU. The Adam optimizer is adopted with an initial learning rate of $2\times10^{-4}$, which is gradually reduced to $2\times10^{-6}$ via a cosine annealing strategy. During training, standard data augmentation techniques, including random rotation and flipping, are applied. The model is trained for 400 epochs with a batch size of 8.

\begin{figure*}[ht]
\centering
\includegraphics[width=\textwidth]{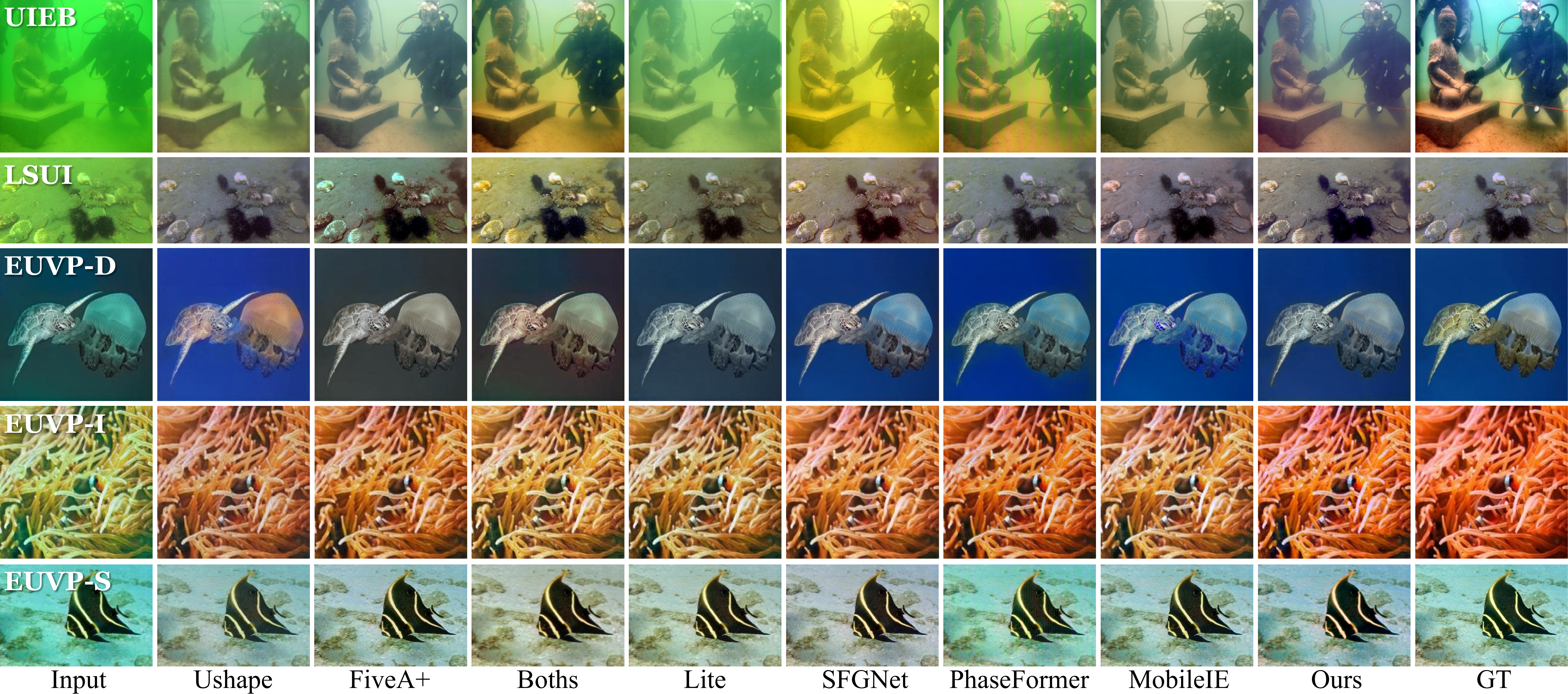}
\caption{Comparison of visual performance across different methods on five full-reference benchmark datasets.}
\vspace{-1em}
\label{06_compare1}
\end{figure*}

\begin{table}[t]
\centering
\fontsize{6pt}{7.5pt}\selectfont
\caption{UCIQE$\uparrow$ performance across different training (UIEB\&LSUI) and validation (U45\&RUIE) sets. Best results are in \textbf{bold}.}
\label{tab:generalization}
\resizebox{0.9\columnwidth}{!}{
\begin{tabular}{lcccccc}
\toprule[0.5pt]
\multirow{2}{*}{Method} & \multicolumn{2}{c}{UIEB} & \multicolumn{2}{c}{LSUI} \\
\cmidrule(lr){2-3} \cmidrule(lr){4-5}
 &U45 & RUIE & U45 & RUIE \\
\midrule[0.3pt]
WaterNet\cite{waternet} & 0.5480 & 0.5154 & 0.5315 & 0.5116 \\
U-shape\cite{U} & 0.5964 & 0.5529 & 0.5654 & 0.5489 \\
UIE-DM\cite{UIEdm} & 0.5854 & 0.5430 & 0.5827 & \textbf{0.5700} \\
SFGNet\cite{sfg} & 0.5937 & 0.5546 & 0.5799 & 0.5564 \\
Lite\cite{lite} & 0.5748 & 0.5396 & 0.5531 & 0.5424 \\
MobileIE\cite{MobileIE} & 0.5840 & 0.5551 & 0.5698 & 0.5546 \\
\midrule[0.3pt]
Ours & \textbf{0.6023} & \textbf{0.5646} & \textbf{0.5880} & 0.5541 \\
\bottomrule[0.5pt]
\end{tabular}
}
\end{table}

\subsubsection{Quantitative Results}
Experimental results are summarized in Table~\ref{tab:main_results}. Across five paired datasets, our method maintains optimal or suboptimal performance evaluated using PSNR, SSIM, and LPIPS. For example, on EUVP-S, our method achieves a PSNR of 27.435 dB and an LPIPS of 0.126. On EUVP-I, it attains the best PSNR and SSIM scores while also obtaining the lowest LPIPS. These results verify the strong robustness of our method under diverse degradation distributions.

To assess cross-domain generalization, we perform inference on the U45 and RUIE datasets using weights trained on UIEB and LSUI datasets, respectively. The results are reported in Table~\ref{tab:generalization}. In the absence of ground-truth, we utilize the UCIQE metric for evaluation. Our method demonstrates stable performance across four settings, achieving the highest scores in UIEB→U45 (0.6023), UIEB→RUIE (0.5646), and LSUI→U45 (0.5880). This indicates that the model preserves excellent contrast and clarity under varying water types and imaging conditions.

\begin{table}[t]
\centering
\caption{CIEDE2000$\downarrow$ values across different camera models. Best results are in \textbf{bold}. (A) U-Shape\cite{U}, (B) Lite\cite{lite}, (C) Boths\cite{boths}, (D) Phaseformer\cite{Phaseformer}, (E) SFGNet\cite{sfg}, (F) MobileIE\cite{MobileIE}.}
\label{tab:ciede2000}
\renewcommand{\arraystretch}{1.1}
\resizebox{\columnwidth}{!}{
\begin{tabular}{lccccccc}
\toprule
Camera & (A) & (B) & (C) & (D) & (E) & (F) & Ours \\
\midrule
Canon D10 & 9.96 & 9.62 & 7.41 & 7.01 & 6.43 & 6.37 & \textbf{5.89} \\
Fujifilm Z33 & 8.23 & 12.01 & 14.03 & 8.00 & \textbf{6.25} & 10.89 & 10.77 \\
Olympus T6000 & 7.71 & 10.49 & 11.12 & \textbf{7.40} & 12.97 & 10.10 & 7.62 \\
Olympus T8000 & 11.14 & \textbf{6.90} & 8.48 & 10.43 & 9.53 & 8.34 & 9.03 \\
Panasonic TS1 & \textbf{9.81} & 12.01 & 10.89 & 17.52 & 14.41 & 12.29 & 13.95 \\
Pentax W60 & 7.87 & 8.19 & 7.10 & 8.63 & 7.15 & 6.70 & \textbf{6.50} \\
Pentax W80 & 9.70 & 7.77 & 7.21 & 8.90 & 8.32 & 7.34 & \textbf{7.11} \\
\midrule
Avg & 9.20 & 9.57 & 9.46 & 9.70 & 9.29 & 8.86 & \textbf{8.70} \\
\bottomrule
\end{tabular}
}
\end{table}

To further evaluate authentic color recovery, we use CIEDE2000 metric on the ColorCheck7 dataset. As shown in Table~\ref{tab:ciede2000}, our method achieves an average color difference of 8.70, the lowest among all compared methods, and yields optimal or competitive performance across multiple camera settings. This validates that our model design effectively restores true underwater colors rather than only boosting contrast or brightness.

Finally, we evaluate model complexity and inference efficiency. Our model is exceptionally lightweight, containing only 3,880 parameters, 0.01MB model size, and 0.145G FLOPs. At $640 \times 480$ resolution, it achieves an inference speed of 409.09 FPS on GPU.

\begin{figure}[!t]
\centering
\includegraphics[width=\columnwidth]{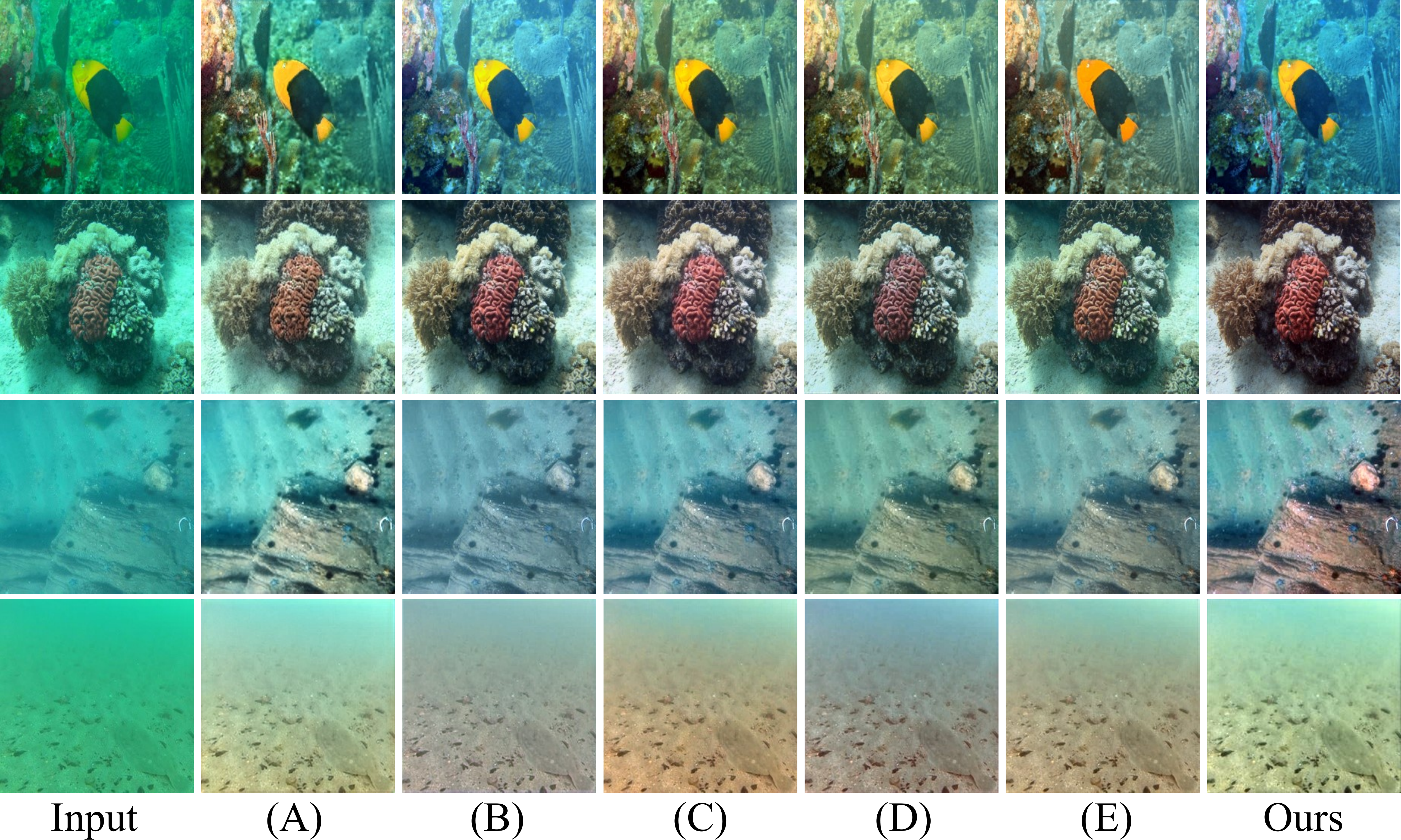}  
\caption{Visual comparison results on U45 and RUIE. From top to bottom, the rows represent inference configurations (Training $\rightarrow$ Testing): (1) UIEB $\rightarrow$ U45, (2) LSUI $\rightarrow$ U45, (3) UIEB $\rightarrow$ RUIE, and (4) LSUI $\rightarrow$ RUIE. Method: (A) U-shape, (B) WaterNet, (C) SFGNet, (D) Lite, and (E) MobileIE (please zoom in for better visualization).
}
\label{07_compare2}
\end{figure}

\begin{figure}[t]
\centering
\includegraphics[width=\columnwidth]{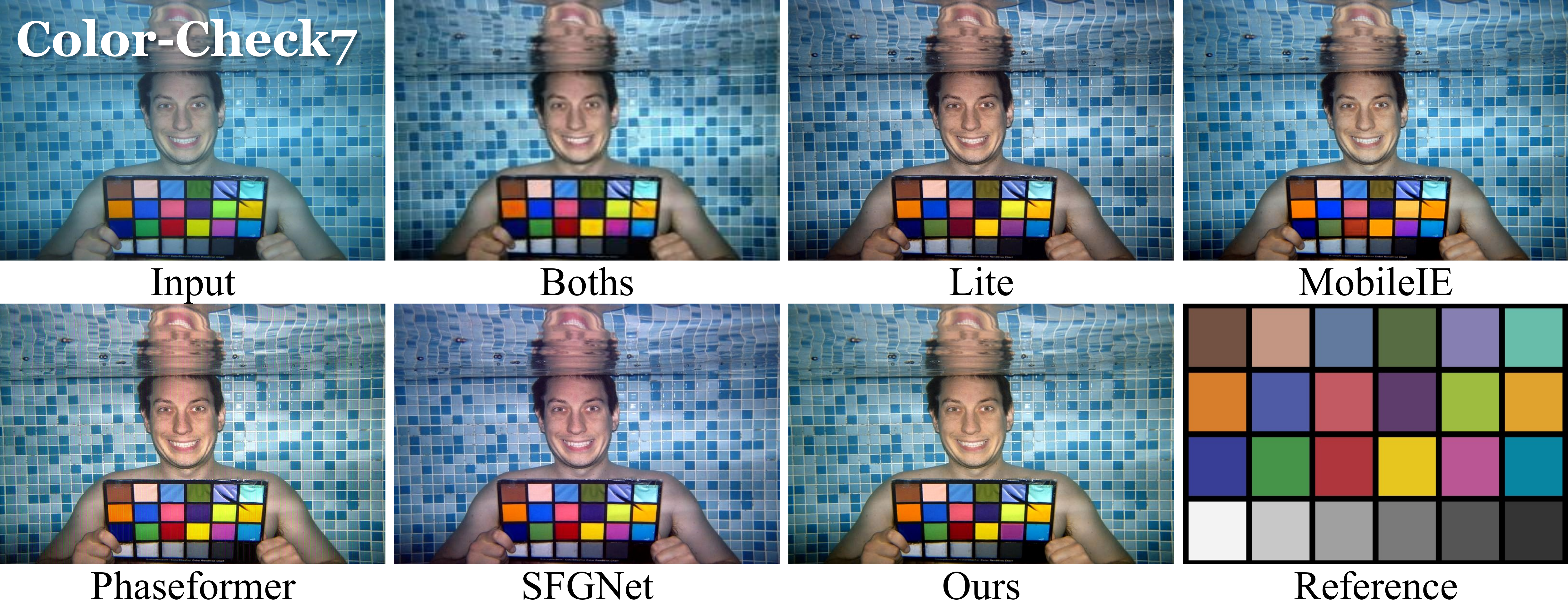} 
\caption{Qualitative results on diverse benchmarks. The bottom-right corners display the reference color checker for Colorcheck7 dataset.}
\label{08_compare3}
\end{figure}

\subsubsection{Qualitative Results}
Visual comparisons across all datasets are shown in Fig.~\ref{06_compare1}. Overall, whereas existing methods often fail to balance color authenticity, structural clarity, and luminance stability, our method provides better visual performance.

In paired datasets, lightweight methods often suffer from over-warming or color distortion during red-channel restoration. Conversely, heavy models (e.g., Phaseformer) may produce texture blurring or edge artifacts in complex environments. Our method restores a more natural and consistent color distribution, avoiding over-correction while reconstructing authentic object colors. Visual comparisons on the U45, RUIE, and ColorCheck7 datasets are presented in Figs.~\ref{07_compare2} and~\ref{08_compare3}, respectively. The results demonstrate that our approach is robust against color distortion, maintaining high visual quality across various underwater scenes.

\subsection{Ablation Study}

\subsubsection{Effectiveness of Proposed Modules}
To validate the effectiveness of each module in our framework, we perform an ablation study using weights trained on the UIEB dataset. Table~\ref{ab1} shows that the full model achieves better performance than the baseline and other variants in both PSNR and SSIM. In particular, the PSNR is improved by approximately 1.828 dB over the baseline, which proves that the three modules work together effectively to improve image quality. The results demonstrate that performance declines when any single module is removed. These results suggest that MRDConv contributes to structural reconstruction, while AWCC and SGCA mainly improve color restoration.

\begin{table}[t]
\centering
\caption{Ablation analysis of the three components.}
\label{ab1}
\resizebox{\columnwidth}{!}{
\begin{tabular}{lccccc}
\toprule
Method & AWCC & MRDConv & SGCA & PSNR$\uparrow$ & SSIM$\uparrow$ \\
\midrule
Baseline &  &  &  & 22.227 & 0.891 \\
(A)    &  & $\checkmark$ & $\checkmark$ & 22.518 & 0.881 \\
(B)   & $\checkmark$ &  & $\checkmark$ & 23.686 & 0.886 \\
(C)   & $\checkmark$ & $\checkmark$ &  & 23.370 & 0.894 \\
\midrule
Full     & $\checkmark$ & $\checkmark$ & $\checkmark$ & \textbf{24.055} & \textbf{0.902} \\
\bottomrule
\end{tabular}
}
\end{table}

\begin{figure}[t]
\centering
\includegraphics[width=0.98\columnwidth]{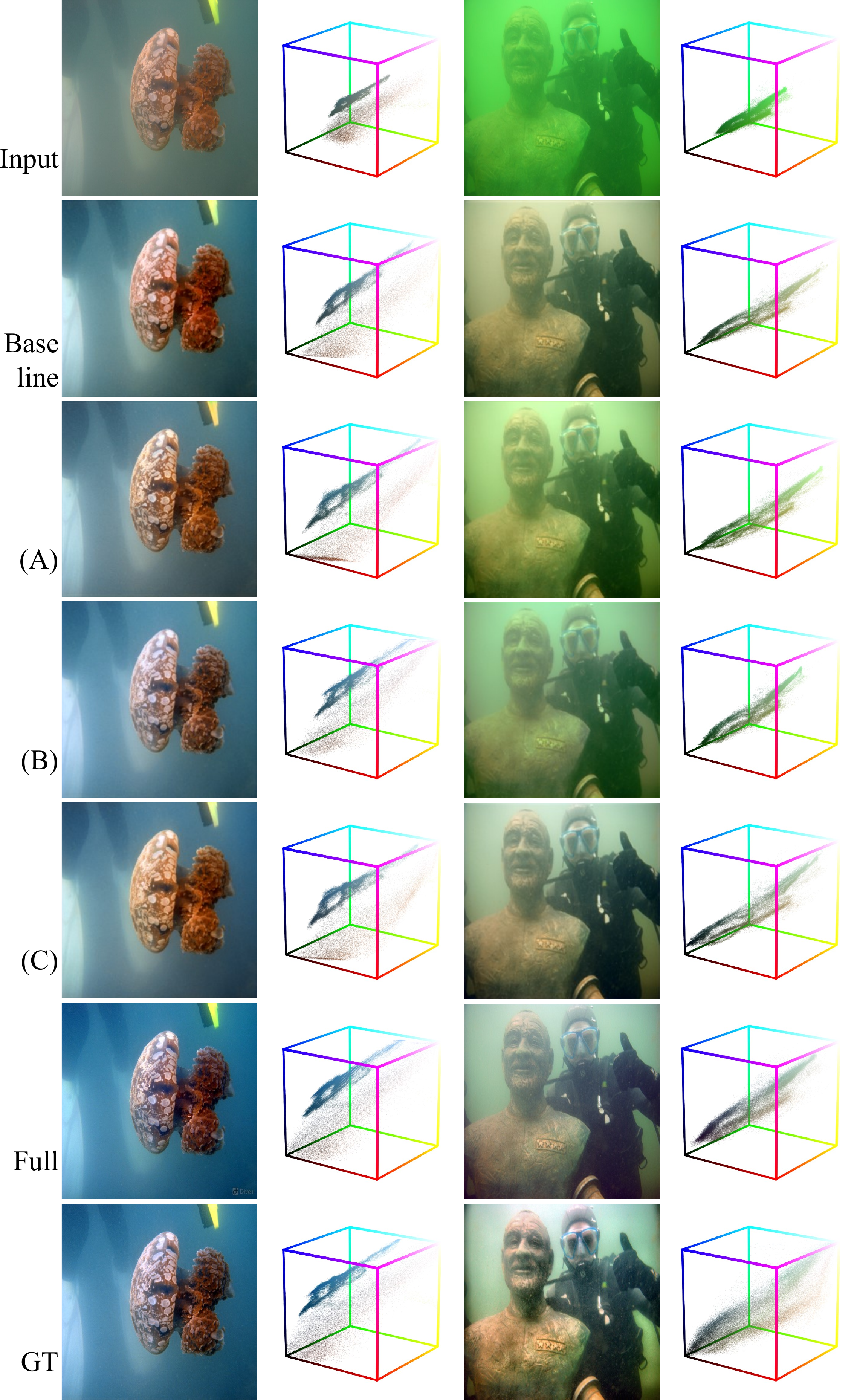}
\caption{Visual ablation results of the proposed modules. The bottom row displays the corresponding sRGB color space point cloud distributions, illustrating the progressive correction of color degradation.}
\label{03_ab1}
\end{figure}

Visual results further confirm these findings. Fig.~\ref{03_ab1} shows that the original input images suffer from a severe green shift. However, as each module is added, the contrast, saturation, and texture clarity gradually move closer to ground-truth. The sRGB space point cloud distribution maps clearly illustrate the color restoration process. In the input, the color distribution is extremely narrow. In contrast, our full model is the most similar to the ground-truth. This demonstrates the proposed method effectively fixes underwater color shifts and achieves high-quality visual restoration.

\subsubsection{Effectiveness of SGCA}

To verify the effectiveness of each component in SGCA, we perform an ablation study by excluding individual elements from the feature vector $V_{stat}$ using weights trained with UIEB dataset.

As reported in Table~\ref{tab:ablation2}, removing the channel-wise mean $\mu_c$ (A) degrades PSNR, SSIM, and LPIPS, indicating its importance for correcting global color bias. Excluding the standard deviation $\sigma_c$ (B) also reduces performance, suggesting that it helps preserve global contrast and structural consistency. Removing the bright-region statistic $v_{bright}$ (C) causes clear drops in PSNR and SSIM, although UIQM increases to 3.090, implying that this term benefits fidelity-oriented restoration while UIQM may favor stronger perceptual enhancement. In contrast, removing the dark-region statistic $v_{dark}$ (D) yields the lowest PSNR and one of the highest LPIPS values, confirming its role in modeling backscattering and haze-related degradation.

\begin{figure}
\centering
\includegraphics[width=\columnwidth]{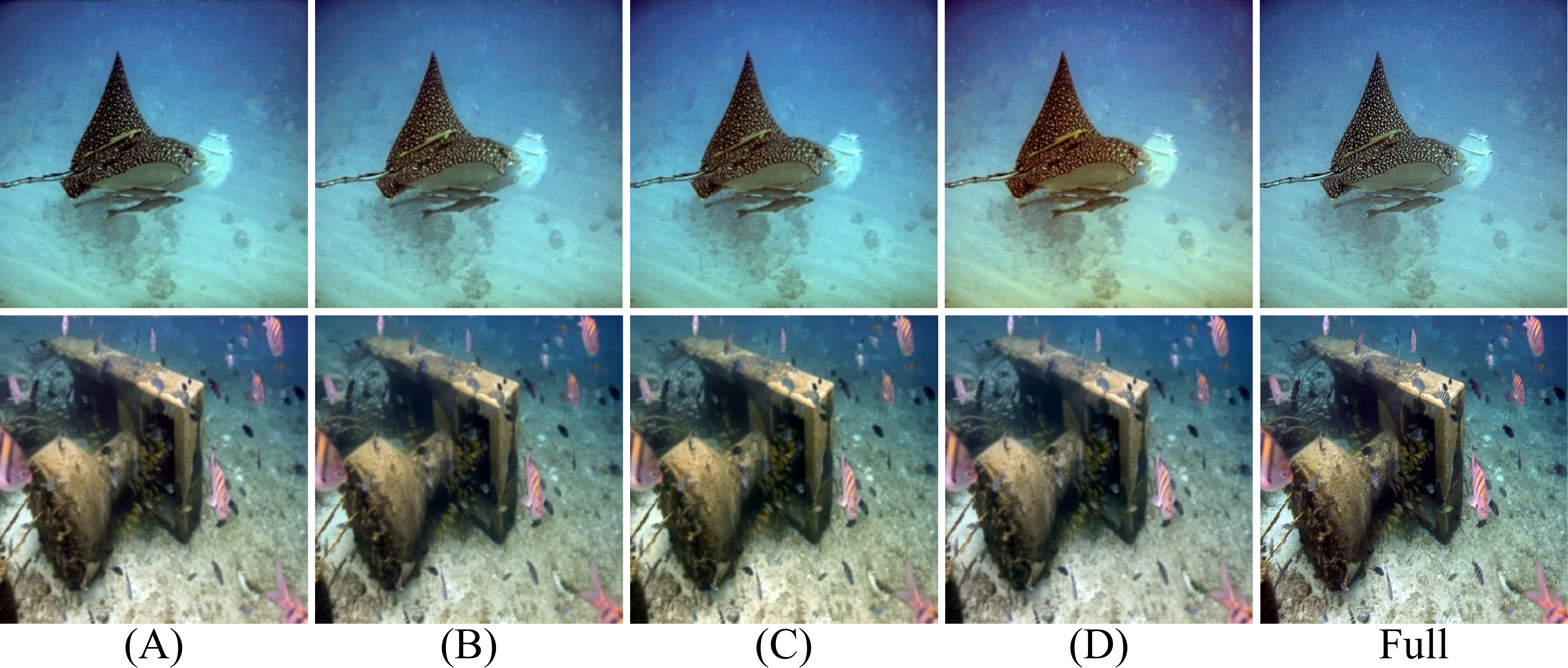}
\caption{Visual comparison of different components in the SGCA module.}
\label{04_ab2}
\end{figure}

As further illustrated in Fig.~\ref{04_ab2}, the full model achieves the best overall balance between fidelity and perceptual quality, demonstrating that these four statistical features are highly complementary. These results verify that SGCA is an effective and interpretable component for underwater color restoration.
\begin{table}[t]
\centering
\caption{Ablation study of different components in SGCA.}
\label{tab:ablation2}
\resizebox{\columnwidth}{!}{
\begin{tabular}{lcccccccc}
\toprule
Method & $\mu_c$ & $\sigma_c$ & $v_{bright}$ & $v_{dark}$

& PSNR$\uparrow$ & SSIM$\uparrow$ & LPIPS$\downarrow$ & UIQM$\uparrow$ \\
\midrule
(A)  &  & $\checkmark$ & $\checkmark$ & $\checkmark$ 
& 24.034 & 0.898 & 0.123 & 3.055 \\

(B)  & $\checkmark$ &  & $\checkmark$ & $\checkmark$ 
& 24.007 & 0.899 & 0.122 & 3.078 \\

(C)  & $\checkmark$ & $\checkmark$ &  & $\checkmark$ 
& 23.721 & 0.894 & 0.128 & \textbf{3.090} \\

(D)  & $\checkmark$ & $\checkmark$ & $\checkmark$ &  
& 23.411 & 0.891 & 0.128 & 3.029 \\
\midrule
Full & $\checkmark$ & $\checkmark$ & $\checkmark$ & $\checkmark$ 
& \textbf{24.055} & \textbf{0.902} & \textbf{0.121} & 3.045 \\
\bottomrule
\end{tabular}
}
\end{table}

\subsubsection{Effectiveness of Loss Functions}
To assess the impact of different loss functions on enhancement quality, we conduct an ablation study using the model trained on UIEB. As presented in Table~\ref{tab:ablation3}, when trained solely with the baseline $L_{char}$ loss, the model achieves a PSNR of 23.133 and an SSIM of 0.891. In contrast, the full model delivers the best results on all evaluation metrics, with PSNR increasing to 24.055 and SSIM reaching 0.902. These findings suggest that the removal of any individual loss term degrades the overall performance.

\begin{figure}[!t]
\centering
\includegraphics[width=\columnwidth]{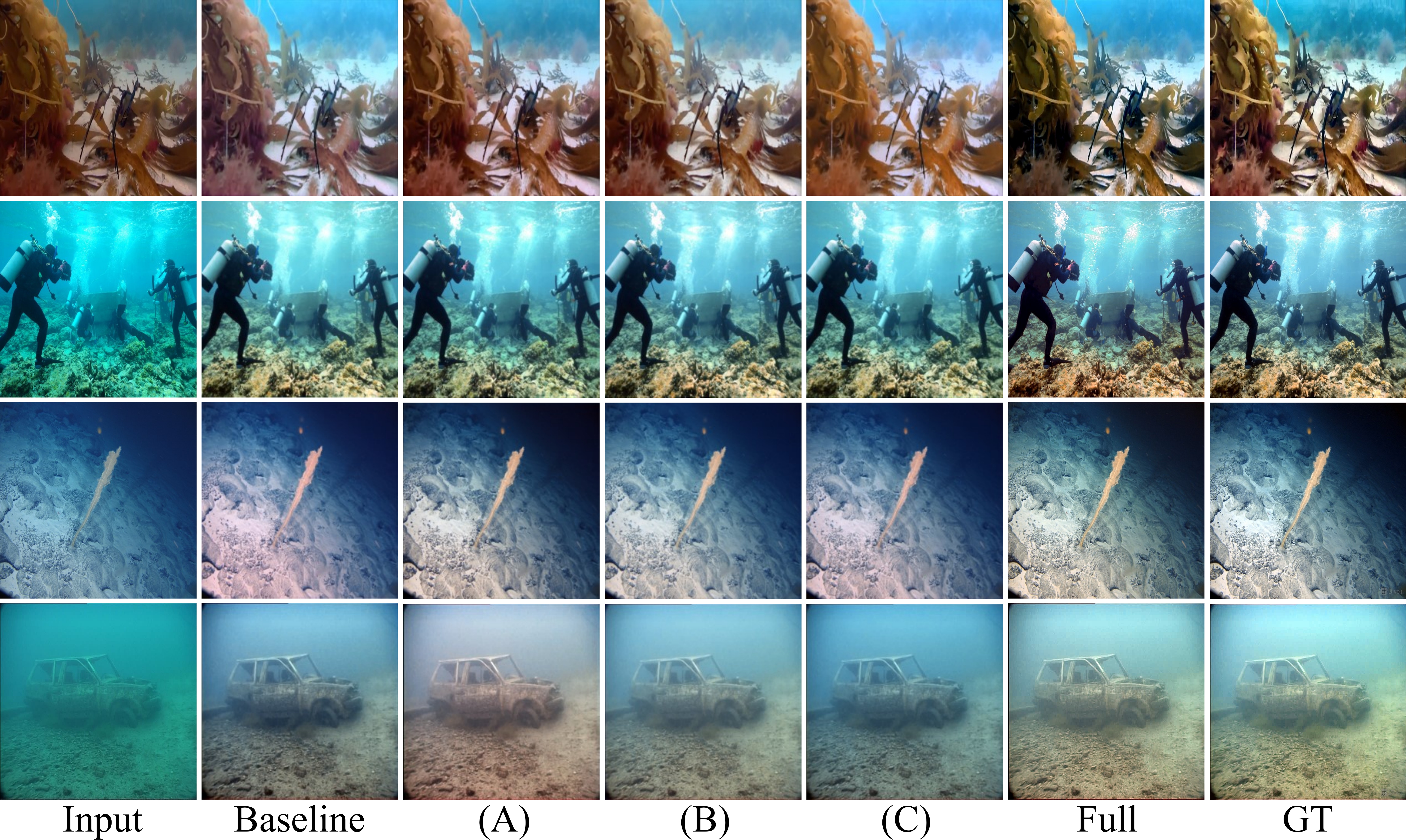}
\caption{Visual ablation results of different loss functions (please zoom in for better visualization).}
\label{05_ab3}
\end{figure}

Specifically, excluding $L_{color}$ results in a noticeable PSNR drop to 24.014. As shown in Fig.~\ref{05_ab3}, this manifests as inaccurate color reproduction, typically characterized by cooler tones and reduced contrast. Furthermore, although $L_{psnr}$ contributes only marginal improvements to quantitative metrics, the joint optimization of all loss components produces enhancement results with superior color consistency and closer visual alignment with the ground-truth.

\begin{table}[t]
\centering
\caption{Ablation study of different loss components.}
\label{tab:ablation3}
\resizebox{\columnwidth}{!}{
\begin{tabular}{lcccccc}
\toprule
Method & $\mathcal{L}_{char}$ & $\mathcal{L}_{psnr}$ & $\mathcal{L}_{vgg}$ & $\mathcal{L}_{color}$ & PSNR$\uparrow$ & SSIM$\uparrow$ \\
\midrule
Baseline & $\checkmark$ &  &  &  & 23.133 & 0.891 \\
(A)    & $\checkmark$ &  & $\checkmark$ & $\checkmark$ & 23.522 & 0.890 \\
(B)    & $\checkmark$ & $\checkmark$ &  & $\checkmark$ & 23.961 & 0.890 \\
(C)    & $\checkmark$ & $\checkmark$ & $\checkmark$ &  & 24.014 & 0.886 \\
\midrule
Full     & $\checkmark$ & $\checkmark$ & $\checkmark$ & $\checkmark$ & \textbf{24.055} & \textbf{0.902} \\
\bottomrule
\end{tabular}
}
\end{table}

\subsubsection{Effectiveness of Rep-scale in MRDConv}
To evaluate the influence of the channel expansion ratio in the MRDConv, we compare two configurations: Rep-scale=4 and Rep-scale=8.

As shown in Table~\ref{tab:repscale}, increasing the Rep-scale from 4 to 8 provides only marginal gains. Rep-scale=8 achieves 24.406 dB PSNR and 0.904 SSIM, improving over Rep-scale=4 by merely 0.351 dB and 0.002, respectively. However, this slight improvement incurs substantial computational overhead. The parameter count increases from 3,880 to 10,870, and FLOPs rise from 0.145G to 0.686G. Meanwhile, inference speed drops from 409.09 FPS to 326.15 FPS, resulting in approximately a 20\% reduction in speed.

These results show clear diminishing returns. Considering the strict efficiency requirements of underwater real-time applications, the increased complexity of Rep-scale=8 is not justified by its limited performance gain. Therefore, Rep-scale=4 is adopted as the default setting, achieving a better balance between restoration quality and speed.

\begin{table}[t]
\centering
\caption{Efficiency and performance analysis for MRDConv.}
\label{tab:repscale}
\resizebox{\linewidth}{!}{
\begin{tabular}{lccccc}
\toprule
\multirow{2}{*}{Method} & Params & FLOPs & FPS & \multirow{2}{*}{PSNR$\uparrow$} & \multirow{2}{*}{SSIM$\uparrow$} \\
\cmidrule(lr){2-4} 
& (K) & (G) & $640\times480$ \\
\midrule
Rep-scale=4   & 3.88     & 0.145  & 409.09   & 24.055     & 0.902      \\
Rep-scale=8   & 10.87    & 0.686    & 326.15 & 24.406     & 0.904      \\ 
\bottomrule
\end{tabular}
}
\end{table}

\begin{figure*}[!t]
\centering
\includegraphics[width=\textwidth]{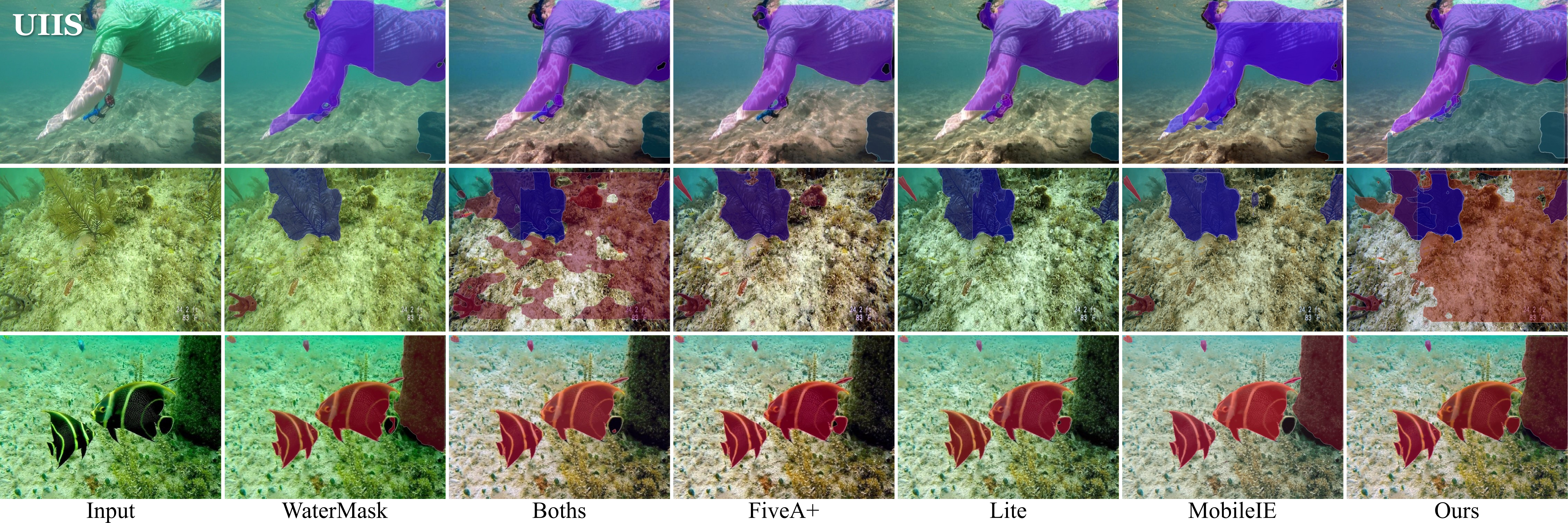}
\caption{Instance segmentation results on the UIIS dataset. Compared with other methods, our model not only enhances overall visual quality, but also preserves sharper boundaries and finer structural details, producing more accurate mask predictions for categories such as Human Divers and Aquatic Plants.}
\label{11_UIIS}
\vspace{-1em}
\end{figure*}

\subsection{High-level Underwater Task}
\subsubsection{Settings}
To validate the effectiveness of the proposed method for downstream underwater tasks, we conduct underwater instance segmentation experiments using the MMDetection framework with an NVIDIA A100 GPU. The UIIS dataset is used, where images are enhanced using different UIE models. The dataset is divided into training and validation subsets with an 8:2 ratio. mAP is used as evaluation metric. The model is trained with the SGD optimizer, with the initial learning rate set to $2.5\times10^{-4}$, momentum set to 0.9, and weight decay set to $1\times10^{-4}$. Standard data augmentation strategies are applied, and training is performed for 12 epochs with a batch size of 16.

\subsubsection{Quantitative Results}
Table~\ref{tab:underwater_map} reports the quantitative analysis of the downstream task evaluation. Overall, our method achieves the highest mAP, surpassing the runner-up methods, FiveA+ and Boths, by +0.003 and +0.004, respectively. This improvement indicates that the proposed enhancement algorithm provides a more robust overall contribution across multiple categories.

A further category-wise analysis reveals that our method achieves peak performance in challenging categories, such as Reefs, Aquatic Plants, and Sea-floor. While existing methods struggle to obtain satisfactory results due to pervasive low contrast and severe blurring in these environments, our approach consistently reaches the highest accuracy. This demonstrates that the proposed algorithm effectively restores critical structural details and enhances object discriminability, thereby boosting the performance of the downstream instance segmentation model.

\begin{table}[t]
\centering
\caption{Instance segmentation performance (mAP) across different enhancement methods on the UIIS dataset.}
\label{tab:underwater_map}
\resizebox{\columnwidth}{!}{%
\begin{tabular}{lcccccccc}
\toprule
Method & Fish & Reefs & \makecell{Aquatic \\ Plants} & \makecell{Wrecks\& \\ Ruins} & \makecell{Human \\ Divers} & Robots & Sea-floor & mAP$\uparrow$ \\
\midrule
WaterMask\cite{watermask}       & 0.442 & 0.260 & 0.076 & 0.184 & 0.473 & 0.075 & 0.066 & 0.225 \\
Boths\cite{boths}           & 0.438 & 0.262 & 0.074 & 0.215 & 0.442 & 0.134 & 0.065 & 0.233 \\
FiveA+\cite{5a}          & 0.434 & 0.256 & 0.078 & 0.192 & 0.478 & 0.143 & 0.054 & 0.234 \\
Lite\cite{lite}  & 0.432 & 0.261 & 0.074 & 0.192 & 0.458 & 0.120 & 0.063 & 0.229 \\
MobileIE \cite{MobileIE}       & 0.438 & 0.254 & 0.072 & 0.184 & 0.472 & 0.138 & 0.060 & 0.231 \\
\midrule
Ours            & 0.440 & \textbf{0.263} & \textbf{0.079} & 0.193 & \textbf{0.478} & 0.135 & \textbf{0.067} & \textbf{0.237} \\
\bottomrule
\end{tabular}
}
\end{table}

\subsubsection{Qualitative Results}

The visual results in Fig.~\ref{11_UIIS} further indicate that the proposed method performs better in scenes with complex textures and overlapping color distributions. Existing methods often fail to clearly separate the foreground from the background and tend to produce discontinuous masks or missing boundary regions. In contrast, our method yields cleaner foreground extraction and more complete mask prediction, especially for slender aquatic plants and Sea-floor boundaries. For categories such as Human Divers, the predicted masks also align better with the object contours.

Overall, these results indicate that the proposed enhancement strategy improves not only visibility and visual quality, but also the structural information required for downstream tasks, thereby providing more reliable features in complex underwater environments.

\section{Application}

Beyond benchmark evaluations, we further verify the effectiveness of the proposed model in real-world applications. Specifically, the model is deployed on an NVIDIA Jetson Orin NX (16GB) platform. To optimize inference efficiency, the model is quantized into the FP16 TensorRT format and reaches a real-time inference speed of 30 FPS at a resolution of $640\times480$. We employ the model weights pre-trained on UIEB for inference in all the following scenarios.

\begin{table}[t]
\centering
\caption{Quantitative quality assessment (UCIQE) under varying turbidity and illumination intensity.}
\label{tab:image_quality}
\resizebox{\columnwidth}{!}{%
\begin{tabular}{ccccccccccc}
\toprule
\multirow{2}{*}{\textbf{NTU}} & \multirow{2}{*}{\textbf{Illum.}} & \multicolumn{4}{c}{\textbf{Raw}} & \multicolumn{4}{c}{\textbf{Enhanced}} & \multirow{2}{*}{\textbf{$\Delta\%$}} \\ \cmidrule(lr){3-6} \cmidrule(lr){7-10}
 & & $\sigma_c$ & ${con}_l$ & $\mu_s$ & UCIQE$\uparrow$ & $\sigma_c$ & ${con}_l$ & $\mu_s$ & UCIQE$\uparrow$ \\
\midrule
\multirow{2}{*}{0.55} & \Rmnum{1} & 0.189 & 0.624 & 0.836 & 0.475 & \textbf{0.287} & \textbf{0.926} & \textbf{0.842} & \textbf{0.605} & 27.3 \\
     & \Rmnum{2} & 0.187 & 0.628 & 0.833 & 0.474 & \textbf{0.270} & \textbf{0.949} & \textbf{0.841} & \textbf{0.604} & 27.3 \\
     \midrule
\multirow{2}{*}{0.99} & \Rmnum{1} & 0.162 & 0.545 & 0.815 & 0.435 & \textbf{0.288} & \textbf{0.926} & \textbf{0.852} & \textbf{0.608} & 39.7 \\
     & \Rmnum{2} & 0.161 & 0.557 & 0.814 & 0.438 & \textbf{0.262} & \textbf{0.941} & \textbf{0.849} & \textbf{0.600} & 36.9 \\
     \midrule
\multirow{2}{*}{2.41} & \Rmnum{1} & 0.183 & 0.537 & 0.782 & 0.435 & \textbf{0.279} & \textbf{0.949} & \textbf{0.839} & \textbf{0.607} & 39.7 \\
     & \Rmnum{2} & 0.171 & 0.533 & 0.773 & 0.425 & \textbf{0.294} & \textbf{0.957} & \textbf{0.834} & \textbf{0.615} & 44.6 \\
     \midrule
\multirow{2}{*}{4.00} & \Rmnum{1} & 0.191 & 0.467 & 0.794 & 0.422 & \textbf{0.266} & \textbf{0.808} & \textbf{0.848} & \textbf{0.565} & 33.8 \\
     & \Rmnum{2} & 0.183 & 0.459 & 0.770 & 0.410 & \textbf{0.288} & \textbf{0.878} & \textbf{0.829} & \textbf{0.590} & 43.8 \\
     \midrule
\multirow{2}{*}{5.42} & \Rmnum{1} & 0.201 & 0.580 & 0.774 & 0.453 & \textbf{0.265} & \textbf{0.875} & \textbf{0.849} & \textbf{0.583} & 28.7 \\
     & \Rmnum{2} & 0.204 & 0.463 & 0.771 & 0.421 & \textbf{0.268} & \textbf{0.655} & \textbf{0.844} & \textbf{0.523} & 24.1 \\
     \midrule
\multirow{2}{*}{6.25} & \Rmnum{1} & 0.219 & 0.537 & 0.769 & 0.448 & \textbf{0.251} & \textbf{0.831} & \textbf{0.844} & \textbf{0.563} & 25.6 \\
     & \Rmnum{2} & 0.207 & 0.526 & 0.796 & 0.446 & \textbf{0.285} & \textbf{0.749} & \textbf{0.848} & \textbf{0.557} & 24.9 \\
     \midrule
\multirow{2}{*}{7.67} & \Rmnum{1} & 0.215 & 0.322 & 0.766 & 0.386 & \textbf{0.260} & \textbf{0.482} & \textbf{0.831} & \textbf{0.468} & 21.1 \\
     & \Rmnum{2} & 0.218 & 0.400 & 0.774 & 0.411 & \textbf{0.274} & \textbf{0.565} & \textbf{0.822} & \textbf{0.495} & 20.4 \\
     \midrule
\multirow{2}{*}{8.82} & \Rmnum{1} & 0.236 & 0.333 & 0.760 & 0.398 & \textbf{0.255} & \textbf{0.510} & \textbf{0.827} & \textbf{0.472} & 18.8 \\
     & \Rmnum{2} & 0.228 & 0.326 & 0.761 & 0.392 & \textbf{0.247} & \textbf{0.502} & \textbf{0.828} & \textbf{0.467} & 19.1 \\
     \midrule
\multicolumn{2}{c}{\textbf{Avg}} &0.197 & 0.490 & 0.787 & 0.429 & \textbf{0.271} & \textbf{0.781} & \textbf{0.839} & \textbf{0.558} & 29.7 \\
\bottomrule
\end{tabular}
}
\end{table}

\begin{figure}[ht]
\centering
\includegraphics[width=\columnwidth]{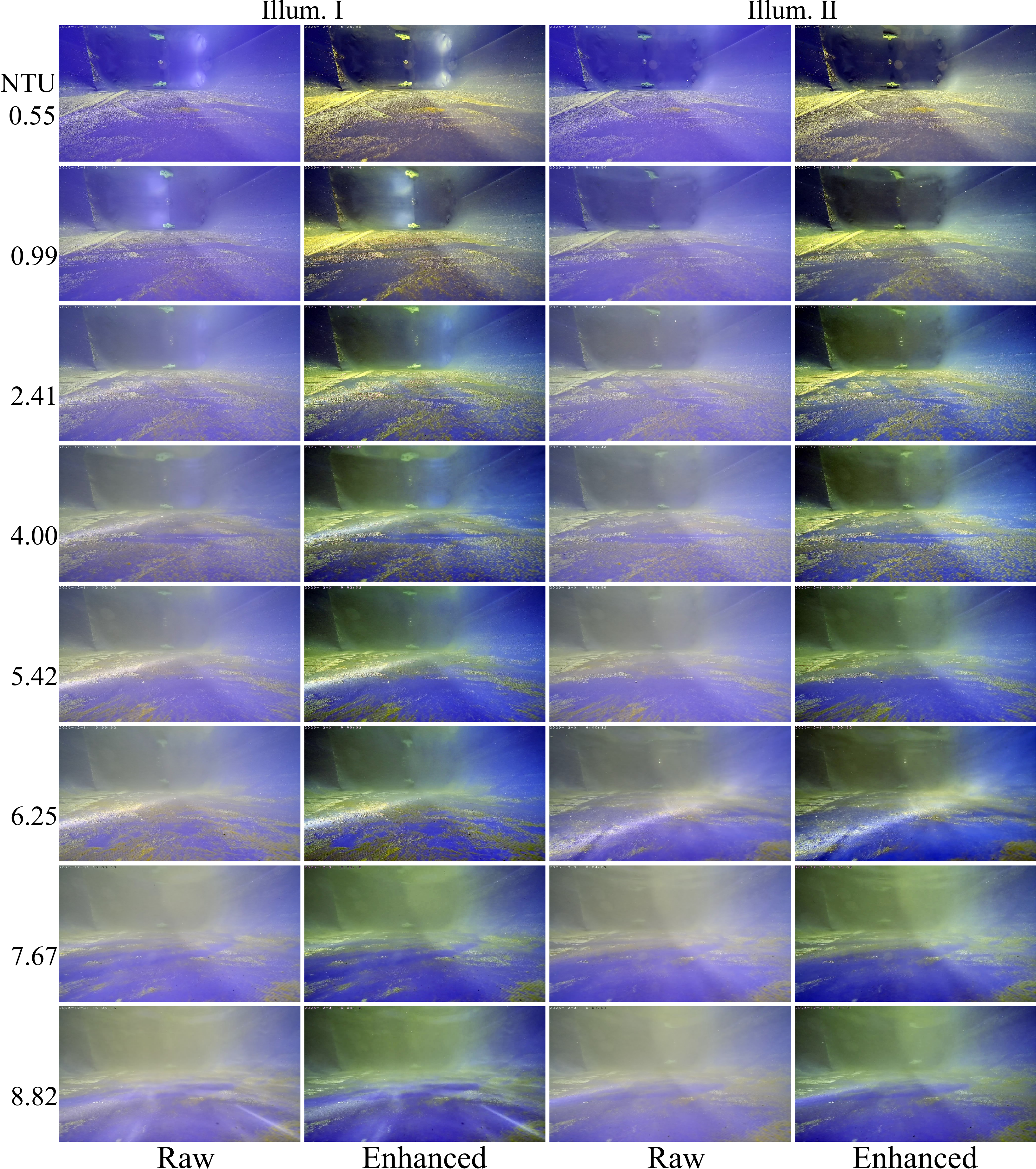}
\caption{Visual results from controlled laboratory tank experiments under varying turbidity levels and illumination intensities, demonstrating the robustness of our method in diverse turbid and lighting conditions.}
\label{09_camera}
\end{figure}

\subsection{Controlled Laboratory Tank}

The controlled experiments were conducted in a laboratory tank lined with blue waterproof fabric to simulate a blue-dominant underwater background. Water turbidity was regulated by adding pond mud, resulting in 16 settings with eight turbidity levels (0.55--8.82 NTU) under two illumination conditions (I: weak light; II: strong light). This range spans from very clear water ($<1$ NTU) to mildly/moderately turbid water (about 4.00–8.82 NTU). In particular, the 2.41 NTU condition is close to that reported for relatively clean reservoir waters such as Qiandao Lake. We integrate our framework on a stereo underwater camera equipped with a zoom lens for testing. 

\begin{equation}
    \mathrm{UCIQE} = 0.4680\times\sigma_c + 0.2745\times{con}_l + 0.2576\times\mu_s
\end{equation}

Image quality is evaluated using UCIQE. Table~\ref{tab:image_quality} shows that the proposed method consistently improved image quality across all settings, increasing the average UCIQE from 0.429 to 0.558, corresponding to a 29.7\% improvement. The largest gain is observed at low-to-moderate turbidity levels (0.99--2.41 NTU), reaching 44.6\%. Even under severe turbidity conditions (7.67--8.82 NTU), the enhanced results still maintained a UCIQE of around 0.47, indicating robust recovery of obscured features. Moreover, the enhanced UCIQE scores under Illumination I and II remained highly consistent; for example, at 0.55 NTU, the scores were 0.605 and 0.604, respectively, demonstrating strong robustness to lighting variations. The corresponding visual results are shown in Fig.~\ref{09_camera}.

\begin{figure}[!t]
\centering
\includegraphics[width=\columnwidth]{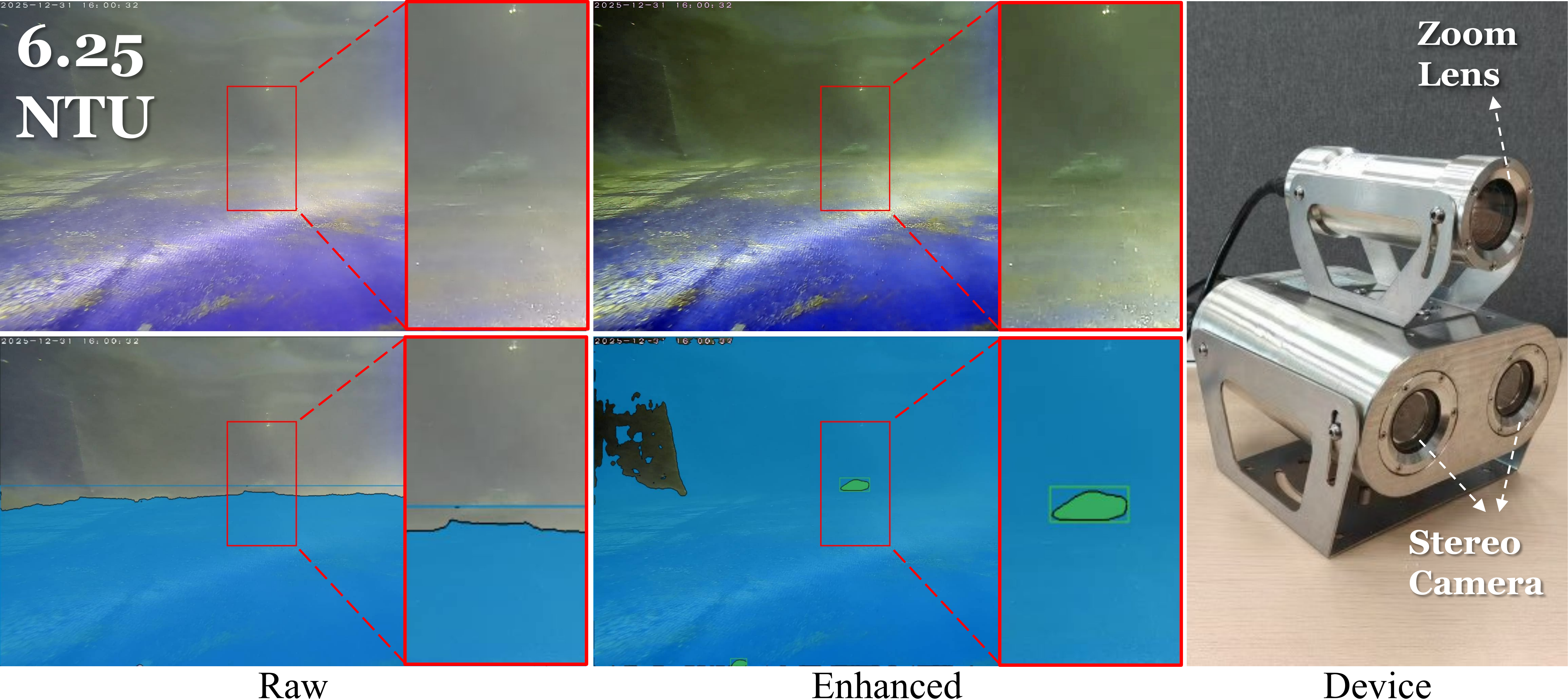}
\caption{Downstream task performance. After enhancement, the model enables precise background segmentation of pool boundaries and successful detection of the toy submarine, which are obscured in the raw frames. For the pool experiments, the stereo camera core is enclosed in housings designed for deep-sea environments.}
\label{12_seg}
\end{figure}

We further evaluated object detection and segmentation on the collected data. As shown in Fig.~\ref{12_seg}, at 6.25 NTU, the raw images suffer from severe degradation and loss of target information, whereas the enhanced results enable clear pool-boundary segmentation and successful identification of the toy submarine. These results verify the effectiveness of our method in improving scene understanding and target perception in turbid underwater environments.

\begin{figure}[ht]
\centering
\includegraphics[width=0.94\columnwidth]{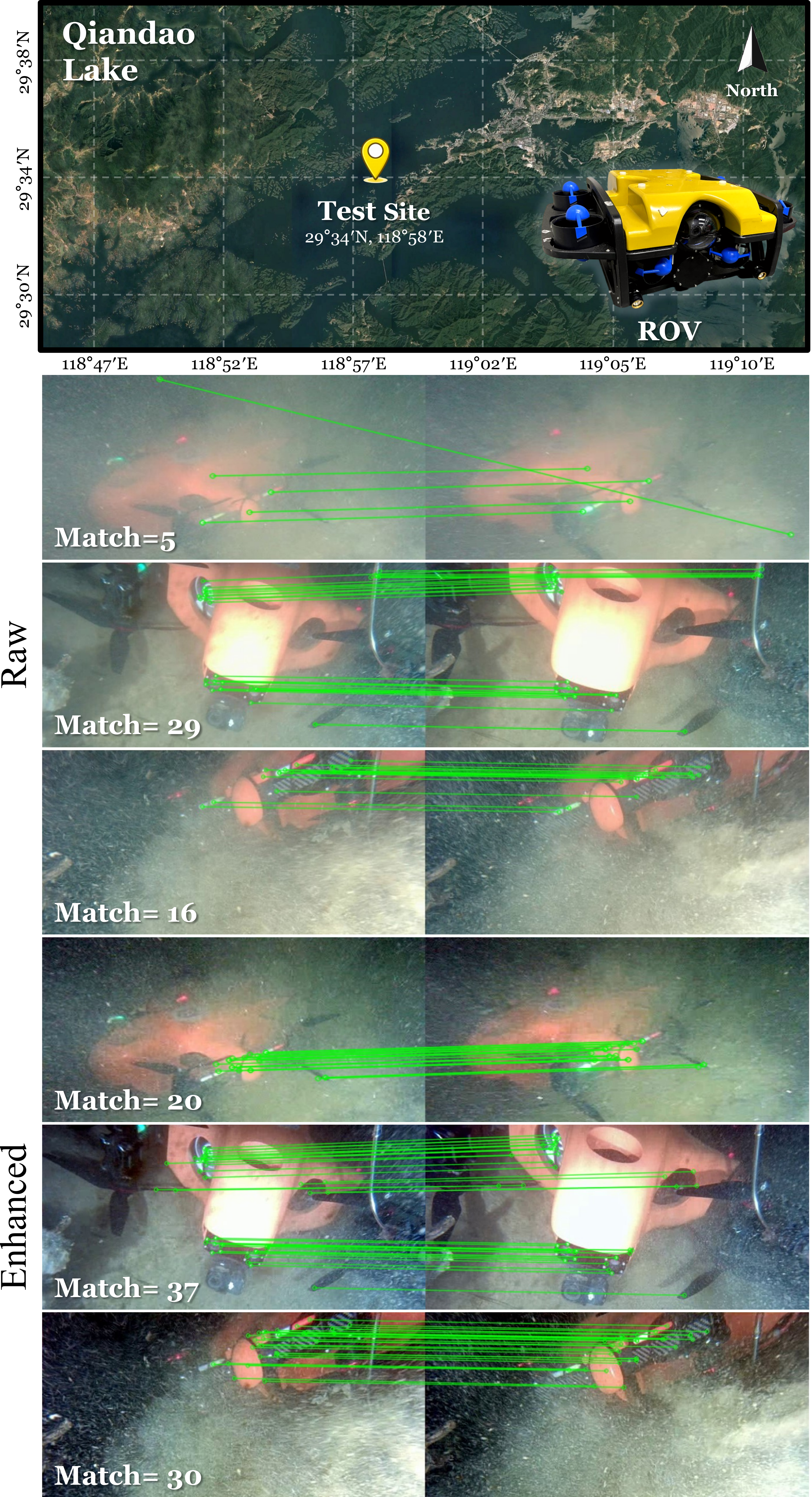}
\caption{Real-world deployment. ROV operation resuspends bottom sediment, causing turbidity and reduced visibility (about 15 m depth, this condition does not correspond to the clear-water state of 2-3 NTU). For the Real-world experiments, the same stereo camera core is used without the deep-sea housings. The SIFT analysis shows that our method significantly increases the number of stable keypoint correspondences between consecutive frames.}
\label{10_rov}
\end{figure}

\subsection{Real-world Deployment on ROV}

To verify the engineering potential of our proposed method in real-world underwater environments, we integrate it into an ROV platform to perform an equipment search-and-recovery mission in Qiandao Lake at a water depth of approximately 15 m. During operation, ROV motion resuspends bottom sediment, resulting in localized turbidity and degraded visual clarity. The onboard camera captures continuous video streams, which are processed in real time on an edge device. To further assess the utility of the enhanced results for visual perception, we conduct SIFT feature matching on consecutive frames. As shown in Fig.~\ref{10_rov}, enhancement significantly increases both the number and stability of valid correspondences. Specifically, the number of successful matches rises from 5 to 20, from 29 to 37, and from 16 to 30 in the three illustrated cases. This improvement indicates that the proposed method restores discriminative structural details and suppresses the adverse effects of turbidity, thereby benefiting inter-frame feature association in challenging underwater scenes. Taken together, these results confirm the robustness and practical value of the proposed method in real-world underwater missions.

\section{Discussion}
The core contribution of this work is achieving an optimal trade-off between color restoration and inference efficiency, a long-standing challenge in underwater vision. While models like DNnet and MobileIE may offer higher throughput, they often face a dilemma between enhancement performance and quality. In contrast, our method maintains better performance, which is beneficial for resource-constrained underwater robots and embedded systems that require high-fidelity perception.

However, a potential bottleneck during hardware deployment is the Top-K operator within the SGCA module. As resolution scales, the computational cost increases rapidly, which may limit ultra-high-definition applications. Future research will focus on optimizing this into a more hardware-friendly operation to unlock even higher real-time performance.

\section{Conclusion}
This paper presented a lightweight underwater image enhancement framework for real-time deployment, achieving an optimal balance between enhancement quality and computational efficiency. Based on the physical properties of underwater imaging, we constructed a pipeline consisting of Adaptive Weighted Channel Compensation, Multi-branch Re-parameterized Dilated Convolution, and Statistical Global Color Adjustment. The results demonstrated that our method achieved state-of-the-art performance across eight datasets while maintaining high inference speed. Further experiments in real-world environments validated its effectiveness in improving both quantitative metrics and downstream task performance. Overall, the proposed method showed strong deployability and could be readily integrated into underwater robotic systems to provide reliable and high-quality visual input for underwater missions.

\vspace{-0.2em}
\bibliographystyle{IEEEtran}
\bibliography{IEEEabrv,UIE}

\begin{thebibliography}{10}
\providecommand{\url}[1]{#1}
\csname url@samestyle\endcsname
\providecommand{\newblock}{\relax}
\providecommand{\bibinfo}[2]{#2}
\providecommand{\BIBentrySTDinterwordspacing}{\spaceskip=0pt\relax}
\providecommand{\BIBentryALTinterwordstretchfactor}{4}
\providecommand{\BIBentryALTinterwordspacing}{\spaceskip=\fontdimen2\font plus
\BIBentryALTinterwordstretchfactor\fontdimen3\font minus \fontdimen4\font\relax}
\providecommand{\BIBforeignlanguage}[2]{{%
\expandafter\ifx\csname l@#1\endcsname\relax
\typeout{** WARNING: IEEEtran.bst: No hyphenation pattern has been}%
\typeout{** loaded for the language `#1'. Using the pattern for}%
\typeout{** the default language instead.}%
\else
\language=\csname l@#1\endcsname
\fi
#2}}
\providecommand{\BIBdecl}{\relax}
\BIBdecl

\bibitem{peng2017underwater}
Y.-T. Peng and P.~C. Cosman, ``Underwater image restoration based on image blurriness and light absorption,'' \emph{IEEE Transactions on Image Processing}, vol.~26, no.~4, pp. 1579--1594, 2017.

\bibitem{PhotoniX}
Z.~Sun, T.~Tian, H.~Hu, Y.~He, M.~Shangguan, T.~Yu, Q.~Yang, M.~Chen, X.~Wang, Y.~Chen \emph{et~al.}, ``Extreme-depth water-related optical imaging: Conquering ultra-low illumination environments from epipelagic zone to {{Mariana Trench}},'' \emph{PhotoniX}, vol.~7, no.~1, p.~7, 2026.

\bibitem{zhang2022underwater}
W.~Zhang, P.~Zhuang, H.-H. Sun, G.~Li, S.~Kwong, and C.~Li, ``Underwater image enhancement via minimal color loss and locally adaptive contrast enhancement,'' \emph{IEEE Transactions on Image Processing}, vol.~31, pp. 3997--4010, 2022.

\bibitem{zongshu01}
S.~Raveendran, M.~D. Patil, and G.~K. Birajdar, ``Underwater image enhancement: A comprehensive review, recent trends, challenges and applications,'' \emph{Artificial Intelligence Review}, vol.~54, no.~7, pp. 5413--5467, 2021.

\bibitem{single}
C.~O. Ancuti and C.~Ancuti, ``Single image dehazing by multi-scale fusion,'' \emph{IEEE Transactions on Image Processing}, vol.~22, no.~8, pp. 3271--3282, 2013.

\bibitem{non_learn}
W.~Zhang, Y.~Wang, and C.~Li, ``Underwater image enhancement by attenuated color channel correction and detail preserved contrast enhancement,'' \emph{IEEE Journal of Oceanic Engineering}, vol.~47, no.~3, pp. 718--735, 2022.

\bibitem{darkchannelpiror}
K.~He, J.~Sun, and X.~Tang, ``Single image haze removal using dark channel prior,'' \emph{IEEE Transactions on Pattern Analysis and Machine Intelligence}, vol.~33, no.~12, pp. 2341--2353, 2010.

\bibitem{transmission}
P.~Drews, E.~Nascimento, F.~Moraes, S.~Botelho, and M.~Campos, ``Transmission estimation in underwater single images,'' in \emph{Proceedings of the {{IEEE}} International Conference on Computer Vision Workshops}, 2013, pp. 825--830.

\bibitem{carlevaris}
N.~{Carlevaris-Bianco}, A.~Mohan, and R.~M. Eustice, ``Initial results in underwater single image dehazing,'' in \emph{Oceans 2010 Mts/{{IEEE}} Seattle}.\hskip 1em plus 0.5em minus 0.4em\relax IEEE, 2010, pp. 1--8.

\bibitem{redchannel}
A.~Galdran, D.~Pardo, A.~Pic{\'o}n, and A.~{Alvarez-Gila}, ``Automatic red-channel underwater image restoration,'' \emph{Journal of Visual Communication and Image Representation}, vol.~26, pp. 132--145, 2015.

\bibitem{support1}
S.~Serikawa and H.~Lu, ``Underwater image dehazing using joint trilateral filter,'' \emph{Computers \& Electrical Engineering}, vol.~40, no.~1, pp. 41--50, 2014.

\bibitem{redbuchang}
C.~O. Ancuti, C.~Ancuti, C.~De~Vleeschouwer, and P.~Bekaert, ``Color balance and fusion for underwater image enhancement,'' \emph{IEEE Transactions on Image Processing}, vol.~27, no.~1, pp. 379--393, 2017.

\bibitem{an2024uwmamba}
G.~An, A.~He, Y.~Wang, and J.~Guo, ``Uwmamba: {{Underwater}} image enhancement with state space model,'' \emph{IEEE Signal Processing Letters}, vol.~31, pp. 2725--2729, 2024.

\bibitem{watergan}
J.~Li, K.~A. Skinner, R.~M. Eustice, and M.~{Johnson-Roberson}, ``{{WaterGAN}}: {{Unsupervised}} generative network to enable real-time color correction of monocular underwater images,'' \emph{IEEE Robotics and Automation letters}, vol.~3, no.~1, pp. 387--394, 2017.

\bibitem{tianyufeng}
Y.~Tian, Y.~Chen, Z.~Sun, L.~Chen, M.~Dou, J.~Lu, Y.~Zheng, and X.~Li, ``A generative data framework with authentic supervision for underwater image restoration and enhancement,'' \emph{arXiv preprint arXiv:2511.14521}, 2025.

\bibitem{UIEdm}
Y.~Tang, H.~Kawasaki, and T.~Iwaguchi, ``Underwater image enhancement by transformer-based diffusion model with non-uniform sampling for skip strategy,'' in \emph{Proceedings of the 31st {{ACM}} International Conference on Multimedia}, 2023, pp. 5419--5427.

\bibitem{tcsvt4}
W.~Zhang, L.~Zhou, P.~Zhuang, G.~Li, X.~Pan, W.~Zhao, and C.~Li, ``Underwater image enhancement via weighted wavelet visual perception fusion,'' \emph{IEEE Transactions on Circuits and Systems for Video Technology}, vol.~34, no.~4, pp. 2469--2483, 2023.

\bibitem{prior2}
J.~Y. Chiang and Y.-C. Chen, ``Underwater image enhancement by wavelength compensation and dehazing,'' \emph{IEEE Transactions on Image Processing}, vol.~21, no.~4, pp. 1756--1769, 2011.

\bibitem{prior3}
Y.~Wang, H.~Liu, and L.-P. Chau, ``Single underwater image restoration using adaptive attenuation-curve prior,'' \emph{IEEE Transactions on Circuits and Systems I: Regular Papers}, vol.~65, no.~3, pp. 992--1002, 2017.

\bibitem{lichongyi}
P.~Zhuang, J.~Wu, F.~Porikli, and C.~Li, ``Underwater image enhancement with hyper-laplacian reflectance priors,'' \emph{IEEE Transactions on Image Processing}, vol.~31, pp. 5442--5455, 2022.

\bibitem{UMSHE}
J.~Zhou, L.~Pang, D.~Zhang, and W.~Zhang, ``Underwater image enhancement method via multi-interval subhistogram perspective equalization,'' \emph{IEEE Journal of Oceanic Engineering}, vol.~48, no.~2, pp. 474--488, 2023.

\bibitem{fusion}
C.~Ancuti, C.~O. Ancuti, T.~Haber, and P.~Bekaert, ``Enhancing underwater images and videos by fusion,'' in \emph{2012 {{IEEE}} Conference on Computer Vision and Pattern Recognition}.\hskip 1em plus 0.5em minus 0.4em\relax IEEE, 2012, pp. 81--88.

\bibitem{WFAC}
W.~Zhang, Q.~Liu, H.~Lu, J.~Wang, and J.~Liang, ``Underwater image enhancement via wavelet decomposition fusion of advantage contrast,'' \emph{IEEE Transactions on Circuits and Systems for Video Technology}, 2025.

\bibitem{tcsvt}
X.~Guo, X.~Chen, S.~Wang, and C.-M. Pun, ``Underwater image restoration through a prior guided hybrid sense approach and extensive benchmark analysis,'' \emph{IEEE Transactions on Circuits and Systems for Video Technology}, vol.~35, no.~5, pp. 4784--4800, 2025.

\bibitem{lixuelong03}
X.~Li, H.~An, H.~Zhao, G.~Li, B.~Liu, X.~Wang, G.~Cheng, G.~Wu, and Z.~Sun, ``Streaknet-arch: an anti-scattering network-based architecture for underwater carrier lidar-radar imaging,'' \emph{IEEE Transactions on Image Processing}, 2025.

\bibitem{cnn02}
A.~Naik, A.~Swarnakar, and K.~Mittal, ``Shallow-uwnet: {{Compressed}} model for underwater image enhancement (student abstract),'' in \emph{Proceedings of the {{AAAI}} Conference on Artificial Intelligence}, vol.~35, 2021, pp. 15\,853--15\,854.

\bibitem{smdr-si}
D.~Zhang, J.~Zhou, C.~Guo, W.~Zhang, and C.~Li, ``Synergistic multiscale detail refinement via intrinsic supervision for underwater image enhancement,'' in \emph{Proceedings of the {{AAAI}} Conference on Artificial Intelligence}, vol.~38, 2024, pp. 7033--7041.

\bibitem{UWCNN}
C.~Li, S.~Anwar, and F.~Porikli, ``Underwater scene prior inspired deep underwater image and video enhancement,'' \emph{Pattern Recognition}, vol.~98, p. 107038, 2020.

\bibitem{waternet}
C.~Li, C.~Guo, W.~Ren, R.~Cong, J.~Hou, S.~Kwong, and D.~Tao, ``An underwater image enhancement benchmark dataset and beyond,'' \emph{IEEE Transactions on Image Processing}, vol.~29, pp. 4376--4389, 2019.

\bibitem{wweuie}
C.-H. Cheng, J.-W. Lee, C.-M. Lee, and C.-C. Hsu, ``{WWE-UIE}: A wavelet \& white balance efficient network for underwater image enhancement,'' in \emph{Proceedings of the IEEE/CVF Winter Conference on Applications of Computer Vision}, 2026, pp. 2135--2145.

\bibitem{UWAGA}
Z.~Huang, J.~Li, Z.~Hua, and L.~Fan, ``Underwater image enhancement via adaptive group attention-based multiscale cascade transformer,'' \emph{IEEE Transactions on Instrumentation and Measurement}, vol.~71, pp. 1--18, 2022.

\bibitem{U}
L.~Peng, C.~Zhu, and L.~Bian, ``U-shape transformer for underwater image enhancement,'' \emph{IEEE Transactions on Image Processing}, vol.~32, pp. 3066--3079, 2023.

\bibitem{Spectroformer}
R.~Khan, P.~Mishra, N.~Mehta, S.~S. Phutke, S.~K. Vipparthi, S.~Nandi, and S.~Murala, ``Spectroformer: {{Multi-domain}} query cascaded transformer network for underwater image enhancement,'' in \emph{Proceedings of the {{IEEE}}/{{CVF}} Winter Conference on Applications of Computer Vision}, 2024, pp. 1454--1463.

\bibitem{Phaseformer}
M.~R. Khan, A.~Negi, A.~Kulkarni, S.~S. Phutke, S.~K. Vipparthi, and S.~Murala, ``Phaseformer: {{Phase-based}} attention mechanism for underwater image restoration and beyond,'' in \emph{2025 {{IEEE}}/{{CVF}} Winter Conference on Applications of Computer Vision ({{WACV}})}.\hskip 1em plus 0.5em minus 0.4em\relax IEEE, 2025, pp. 9618--9629.

\bibitem{Lixuelong02}
X.~Li, Y.~Ren, X.~Jin, C.~Lan, X.~Wang, W.~Zeng, X.~Wang, and Z.~Chen, ``Diffusion models for image restoration and enhancement: {{A}} comprehensive survey,'' \emph{International Journal of Computer Vision}, vol. 133, no.~11, pp. 8078--8108, 2025.

\bibitem{wfdiff}
C.~Zhao, W.~Cai, C.~Dong, and C.~Hu, ``Wavelet-based fourier information interaction with frequency diffusion adjustment for underwater image restoration,'' in \emph{Proceedings of the {{IEEE}}/{{CVF}} Conference on Computer Vision and Pattern Recognition}, 2024, pp. 8281--8291.

\bibitem{Mobilenetv3}
A.~Howard, M.~Sandler, G.~Chu, L.-C. Chen, B.~Chen, M.~Tan, W.~Wang, Y.~Zhu, R.~Pang, V.~Vasudevan \emph{et~al.}, ``Searching for mobilenetv3,'' in \emph{Proceedings of the {{IEEE}}/{{CVF}} International Conference on Computer Vision}, 2019, pp. 1314--1324.

\bibitem{shufflenet}
X.~Zhang, X.~Zhou, M.~Lin, and J.~Sun, ``Shufflenet: {{An}} extremely efficient convolutional neural network for mobile devices,'' in \emph{Proceedings of the {{IEEE}} Conference on Computer Vision and Pattern Recognition}, 2018, pp. 6848--6856.

\bibitem{Ghost}
K.~Han, Y.~Wang, Q.~Tian, J.~Guo, C.~Xu, and C.~Xu, ``Ghostnet: {{More}} features from cheap operations,'' in \emph{Proceedings of the {{IEEE}}/{{CVF}} Conference on Computer Vision and Pattern Recognition}, 2020, pp. 1580--1589.

\bibitem{chen2023run}
J.~Chen, S.-h. Kao, H.~He, W.~Zhuo, S.~Wen, C.-H. Lee, and S.-H.~G. Chan, ``Run, don't walk: Chasing higher {{FLOPS}} for faster neural networks,'' in \emph{Proceedings of the {{IEEE}}/{{CVF}} Conference on Computer Vision and Pattern Recognition}, 2023, pp. 12\,021--12\,031.

\bibitem{fastvit}
P.~K.~A. Vasu, J.~Gabriel, J.~Zhu, O.~Tuzel, and A.~Ranjan, ``Fastvit: {{A}} fast hybrid vision transformer using structural reparameterization,'' in \emph{Proceedings of the {{IEEE}}/{{CVF}} International Conference on Computer Vision}, 2023, pp. 5785--5795.

\bibitem{repvit}
A.~Wang, H.~Chen, Z.~Lin, J.~Han, and G.~Ding, ``Repvit: {{Revisiting}} mobile cnn from vit perspective,'' in \emph{Proceedings of the {{IEEE}}/{{CVF}} Conference on Computer Vision and Pattern Recognition}, 2024, pp. 15\,909--15\,920.

\bibitem{MobileIE}
H.~Yan, A.~Li, X.~Zhang, Z.~Liu, Z.~Shi, C.~Zhu, and L.~Zhang, ``Mobile{IE}: {{An}} extremely lightweight and effective convnet for real-time image enhancement on mobile devices,'' in \emph{Proceedings of the {{IEEE}}/{{CVF}} International Conference on Computer Vision}, 2025, pp. 21\,949--21\,960.

\bibitem{cnn01}
M.~J. Islam, Y.~Xia, and J.~Sattar, ``Fast underwater image enhancement for improved visual perception,'' \emph{IEEE Robotics and Automation letters}, vol.~5, no.~2, pp. 3227--3234, 2020.

\bibitem{puie}
Z.~Fu, W.~Wang, Y.~Huang, X.~Ding, and K.-K. Ma, ``Uncertainty inspired underwater image enhancement,'' in \emph{European Conference on Computer Vision}.\hskip 1em plus 0.5em minus 0.4em\relax Springer, 2022, pp. 465--482.

\bibitem{dnnet}
T.~Cao, Z.~Yu, and B.~Zheng, ``{{DNnet}}: {{A}} lightweight network for real-time {{4K}} underwater image enhancement using dynamic range and average normalization,'' \emph{Expert Systems with Applications}, vol. 270, p. 126561, 2025.

\bibitem{5a}
J.~Jiang, T.~Ye, S.~Chen, E.~Chen, Y.~Liu, S.~Jun, J.~Bai, and W.~Chai, ``Five a+ network: You only need 9k parameters for underwater image enhancement,'' in \emph{34th British Machine Vision Conference 2023, {BMVC} 2023, Aberdeen, UK, November 20-24, 2023}.\hskip 1em plus 0.5em minus 0.4em\relax BMVA, 2023.

\bibitem{sfg}
C.~Zhao, W.~Cai, C.~Dong, and Z.~Zeng, ``Toward sufficient spatial-frequency interaction for gradient-aware underwater image enhancement,'' in \emph{{{ICASSP}} 2024-2024 {{IEEE}} International Conference on Acoustics, Speech and Signal Processing ({{ICASSP}})}.\hskip 1em plus 0.5em minus 0.4em\relax IEEE, 2024, pp. 3220--3224.

\bibitem{lite}
S.~Zhang, S.~Zhao, D.~An, D.~Li, and R.~Zhao, ``{{LiteEnhanceNet}}: {{A}} lightweight network for real-time single underwater image enhancement,'' \emph{Expert Systems with Applications}, vol. 240, p. 122546, 2024.

\bibitem{boths}
X.~Liu, S.~Lin, K.~Chi, Z.~Tao, and Y.~Zhao, ``Boths: {{Super}} lightweight network-enabled underwater image enhancement,'' \emph{IEEE Geoscience and Remote Sensing Letters}, vol.~20, pp. 1--5, 2022.

\bibitem{u45}
H.~Li, J.~Li, and W.~Wang, ``A fusion adversarial underwater image enhancement network with a public test dataset,'' \emph{arXiv preprint arXiv:1906.06819}, 2019.

\bibitem{ruie}
R.~Liu, X.~Fan, M.~Zhu, M.~Hou, and Z.~Luo, ``Real-world underwater enhancement: Challenges, benchmarks, and solutions under natural light,'' \emph{IEEE Transactions on Circuits and Systems for Video Technology}, vol.~30, no.~12, pp. 4861--4875, 2020.

\bibitem{quality}
Z.~Wang, A.~C. Bovik, H.~R. Sheikh, and E.~P. Simoncelli, ``Image quality assessment: from error visibility to structural similarity,'' \emph{IEEE Transactions on Image Processing}, vol.~13, no.~4, pp. 600--612, 2004.

\bibitem{Uiqm}
K.~Panetta, C.~Gao, and S.~Agaian, ``Human-visual-system-inspired underwater image quality measures,'' \emph{IEEE Journal of Oceanic Engineering}, vol.~41, no.~3, pp. 541--551, 2015.

\bibitem{Uciqe}
M.~Yang and A.~Sowmya, ``An underwater color image quality evaluation metric,'' \emph{IEEE Transactions on Image Processing}, vol.~24, no.~12, pp. 6062--6071, 2015.

\bibitem{ciede2000}
G.~Sharma, W.~Wu, and E.~N. Dalal, ``The {CIEDE2000} color-difference formula: Implementation notes, supplementary test data, and mathematical observations,'' \emph{Color Res. Appl.}, vol.~30, no.~1, pp. 21--30, Feb.2005.

\bibitem{watermask}
S.~Lian, H.~Li, R.~Cong, S.~Li, W.~Zhang, and S.~Kwong, ``Watermask: {{Instance}} segmentation for underwater imagery,'' in \emph{Proceedings of the {{IEEE}}/{{CVF}} International Conference on Computer Vision}, 2023, pp. 1305--1315.

\end{thebibliography}

\end{document}